\documentclass[tightenlines, twocolumn, notitlepage, nofootinbib, superscriptaddress]{revtex4-2}
\pdfoutput=1

\usepackage{amsmath}
\usepackage{amssymb}
\usepackage{bibunits}
\usepackage{enumerate}
\usepackage{floatrow}
\usepackage{hyperref}
\usepackage[symbol]{footmisc}
\usepackage{graphicx}
\usepackage{units}
\usepackage[dvipsnames]{xcolor}
\usepackage{xspace}

\graphicspath{{./}, {Graphics/}}
\setcitestyle{super}

\newcommand{\vect}[1]{\mathbf{#1}}

\newcommand{\phoenics}{\textsc{Phoenics}\xspace}
\newcommand{\gryffin}{\textsc{Gryffin}\xspace}

\newfloatcommand{capbtabbox}{table}[][\FBwidth]

\makeatletter
\renewcommand*{\p@subsection}{\thesection.}
\makeatother

\begin{document}


	\title{\large{} Gryffin: An algorithm for Bayesian optimization of categorical variables informed by expert knowledge}

	\date{\today}

	\author{Florian H\"ase}
	\email{hase.florian@gmail.com}
	\affiliation{Department of Chemistry and Chemical Biology, Harvard University, Cambridge, Massachusetts, 02138, USA}
	\affiliation{Vector Institute for Artificial Intelligence, Toronto, ON M5S 1M1, Canada}
	\affiliation{Department of Computer Science, University of Toronto, Toronto, ON M5S 3H6, Canada}
	\affiliation{Department of Chemistry, University of Toronto, Toronto, ON M5S 3H6, Canada}
	\author{Matteo Aldeghi}
	\affiliation{Vector Institute for Artificial Intelligence, Toronto, ON M5S 1M1, Canada}
	\affiliation{Department of Computer Science, University of Toronto, Toronto, ON M5S 3H6, Canada}
	\affiliation{Department of Chemistry, University of Toronto, Toronto, ON M5S 3H6, Canada}
	\author{Riley J. Hickman}
	\affiliation{Department of Computer Science, University of Toronto, Toronto, ON M5S 3H6, Canada}
	\affiliation{Department of Chemistry, University of Toronto, Toronto, ON M5S 3H6, Canada}
	\author{Lo\"ic M. Roch}
	\affiliation{Vector Institute for Artificial Intelligence, Toronto, ON M5S 1M1, Canada}
	\affiliation{Department of Computer Science, University of Toronto, Toronto, ON M5S 3H6, Canada}
	\affiliation{Department of Chemistry, University of Toronto, Toronto, ON M5S 3H6, Canada}
	\affiliation{Atinary Technologies S\`arl, 1006 Lausanne, VD, Switzerland}
	\author{Al\'an Aspuru-Guzik}
	\email{alan@aspuru.com}
	\affiliation{Vector Institute for Artificial Intelligence, Toronto, ON M5S 1M1, Canada}
	\affiliation{Department of Computer Science, University of Toronto, Toronto, ON M5S 3H6, Canada}
	\affiliation{Department of Chemistry, University of Toronto, Toronto, ON M5S 3H6, Canada}
	\affiliation{Lebovic Fellow, Canadian Institute for Advanced Research, Toronto, Ontario M5G 1Z8, Canada}

	\begin{abstract}
	
	Designing functional molecules and advanced materials requires complex design choices: tuning continuous process parameters such as temperatures or flow rates, while simultaneously selecting catalysts or solvents. 
	To date, the development of data-driven experiment planning strategies for autonomous experimentation has largely focused on continuous process parameters despite the urge to devise efficient strategies for the selection of categorical variables. 
	Here, we introduce \gryffin, a general purpose optimization framework for the autonomous selection of categorical variables driven by expert knowledge. 
	\gryffin augments Bayesian optimization based on kernel density estimation with smooth approximations to categorical distributions. 
	Leveraging domain knowledge in the form of physicochemical descriptors, \gryffin can significantly accelerate the search for promising molecules and materials. 
	\gryffin can further highlight relevant correlations between the provided descriptors to inspire physical insights and foster scientific intuition. 
	In addition to comprehensive benchmarks, we demonstrate the capabilities and performance of \gryffin on three examples in materials science and chemistry: (i) the discovery of non-fullerene acceptors for organic solar cells, (ii) the design of hybrid organic-inorganic perovskites for light harvesting, and (iii) the identification of ligands and process parameters for Suzuki-Miyaura reactions. 
	Our results suggest that \gryffin, in its simplest form, is competitive with state-of-the-art categorical optimization algorithms. 
	However, when leveraging domain knowledge provided \emph{via} descriptors, \gryffin outperforms other approaches while simultaneously refining this domain knowledge to promote scientific understanding.
	
	\end{abstract}

	\maketitle

\begin{bibunit}[unsrt]

	\section{Introduction}

	The discovery of functional molecules and advanced materials is recognized as one of the fundamental obstacles to the development of emerging and future technologies, which are needed to face immediate challenges in clean energy, sustainability and global health.\cite{Tabor:2018, Correa:2018} 
	To date, scientific discovery across chemistry, materials science and biology has been accelerated by combinatorial high-throughput strategies and automated experimentation equipment.\cite{Senkan:1998_ht, Maier:2007_ht, Koinuma:2004_ht, Macarron:2011_ht, Mennen:2019} 
	Despite remarkable successes with high-throughput approaches,\cite{Troshin:2017, Selekman:2017, Cheng:2015, Collins:2014, McNally:2011} the combinatorial explosion of molecular and materials candidates renders exhaustive evaluations impossible. 
	This limitation can be alleviated by the use of adaptive strategies, which selectively explore the search space and only evaluate the most promising materials candidates.\cite{Moore:2011_chemistry} 
	Autonomous platforms combine such adaptive, data-driven strategies with automated experimentation systems and might constitute a next-generation approach to the advancement of scientific discovery.\cite{Hase:2019, Jensen:2019_autonomy, Coley:2019_autonomy, Roch:2018_sr}

	Recently, data-driven experiment planning has experienced increased attention by the scientific community.
	Eminent examples include the search for antimicrobial peptides,\cite{Yoshida:2018_peptides} the synthesis of organic molecules,\cite{Coley:2019_robot, Zhou:2017_rnn} the development of a process for the synthesis of short polymer fibers,\cite{Li:2017} the discovery and crystallization of polyoxometalates,\cite{Duros:2017_crystals} the discovery of metallic glasses,\cite{Ren:2018_closed_loop} the optimization of carbon dioxide-assisted nanoparticle deposition,\cite{Casciato:2012_closed_loop} the optimization of Mitsunobu and deoxyfluorination reactions\cite{Shields:2021}, and the creation of Bose-Einstein condensates.\cite{Wigley:2016} 
	Motivated by the successes of data-driven experiment planning, the development and deployment of autonomous workflows for scientific discovery, as well as their benefits over conventional experimentation strategies, are being actively explored.\cite{Stein:2019_closed_loop} 
	For example, autonomous platforms have been applied to the optimization of reaction conditions,\cite{Christensen:2020,Bedard:2018,Cortes:2018,Fitzpatrick:2015} the unsupervised growth of carbon nanotubes,\cite{Nikolaev:2016, Maruyama:2017} autonomous synchrotron X-ray characterization,\cite{Noack:2019, Kusne:2014_closed_loop} the discovery of thin-film materials,\cite{Macleod:2019_ada} the synthesis of inorganic photoluminescent quantum dots,\cite{Li:2020} and the discovery of photostable quaternary polymer blends for organic photovoltaics.\cite{Langner:2019_closed_loop}

	Although autonomous experimentation platforms and data-driven experiment planning strategies are on the rise, the aforementioned examples mostly targeted optimization tasks involving continuous process parameters. 
	Yet, scientific discovery in chemistry and materials science typically requires the simultaneous optimization of continuous and categorical variables, such as the selection of a catalyst or solvent, which cannot be targeted efficiently with continuous optimization methods.
	In fact, the iterative optimization of a molecule itself may be framed as a categorical optimization task.
	Approaches for the data-driven selection of categorical parameters are often handcrafted and demand human decisions, which can adversely affect the experimentation throughput and its efficiency.\cite{Shields:2021}
	Examples of experimentation workflows involving the selection of categorical variables with partial human interaction have been demonstrated in the context reaction optimization.\cite{Baumgartner:2018, Reizman:2016, Reizman:2015} 
	More recently, strategies for the data-driven optimization of categorical spaces have started being proposed.\cite{Zhang:2020,Shields:2021} Yet, the limited availability of general-purpose approaches for the efficient optimization of categorical variables in chemistry and materials science is a challenge to autonomous discovery workflows and a major obstacle to the successful deployment of autonomous experimentation platforms.

	The machine learning community has been exploring algorithmic approaches to the data-driven selection of categorical variables in the context of hyperparameter optimization\cite{Snoek:2012_bopt} and robotics\cite{Lizotte:2007}. 
	Yet, these applications are different from optimization tasks in chemistry and materials science, where categorical variables can usually be characterized by a notion of similarity between individual choices.
	In fact, the concept of molecular, or chemical, similarity is well established in chemistry and is central to cheminformatics.\cite{Johnson:1990}
	For example, co-polymers for hydrogen production can be synthesized from monomeric units that may be clustered based on their reactivities.\cite{Bai:2019} 
	More generally, a quantitative measure of similarity between molecules and materials can be defined based on their physical, chemical and structural properties. 
	An experiment planning strategy that actively leverages physical, chemical, or structural descriptors is expected to accelerate scientific discovery, thanks to the exploitation of expert domain knowledge. 
	Furthermore, the ability of these descriptors to predict target properties of interest would allow one to gain scientific insights from optimization campaigns, thus potentially inspiring the design of additional molecules and materials not initially considered in the chemical search space.

	In this work we introduce \gryffin, a global optimization strategy for the selection of categorical variables in chemistry and materials science, which can be readily integrated into autonomous workflows. 
	\gryffin implements a Bayesian optimization framework leveraging kernel density estimation directly on the categorical space, which can be accelerated with domain knowledge in the form of physicochemical descriptors by locally redefining the metric on the categorical space. 
	In addition to accelerating the identification of optimal categorical options via the use of descriptors, \gryffin can construct more informative ones on-the-fly. 
	This feature can highlight the relevance of the provided descriptors, enabling the interpretation of their role in determining the target properties of interest. 
	\gryffin's approach to categorical optimization differs from how Bayesian optimization has been adopted in automatic chemical design. In the latter, categorical input spaces are mapped onto continuous ones with variational autoencoders and using large datasets.\cite{GomezBombarelli:2018,Griffiths:2020,Kusner:2017} Bayesian optimization is then used to search the continuous space for promising molecules or materials. \gryffin is a Bayesian optimization approach that directly searches the categorical space instead, without having to learn a map between categorical and continuous spaces.

	We highlight the applicability and performance of \gryffin on a set of synthetic benchmark functions and three real-world tasks: the discovery of small molecule non-fullerene acceptors for organic solar cells, the discovery of hybrid organic-inorganic perovskites for light harvesting, and the combined selection of ligands and process parameters for the optimization of Suzuki-Miyaura coupling reactions. 
	We identify four key advantages of \gryffin: (i) it provides a framework for categorical optimization that is competitive with state-of-the-art algorithms in its na\"{i}ve formulation, (ii) it outperforms the state-of-the-art when taking advantage of expert knowledge in the form of descriptors, (iii) can inspire scientific insights, and (iv) can be readily integrated with continuous optimization strategies to enable the robust and efficient optimization of mixed continuous-categorical domains in both sequential and batched workflows.

	\section{Background and related work}
\label{sec:background}

	Experiment planning can be formulated as an optimization task, where we consider a set of controllable parameters within a defined domain, $\vect{z} \in \mathcal{Z}^n$, and an experimental response, $f(z)$, for each of the parameter choices. 
	In the context of reaction optimization, the controllable parameter could for example include the reaction temperature, the amount of solvent, or the choice of catalyst, while the experimental response could be the reaction yield, or the rate at which the product is formed. 
	The optimization domain is also referred to as the \emph{design space} or the \emph{candidate space}. 
	The optimization task in experiment planning consists in the identification of specific parameter values, $\vect{z}^* \in \mathcal{Z}^n$, which yield the desired experimental outcome, $f(\vect{z}^*)$. 
	For simplicity, we will consider minimization tasks from here on, i.e. we formulate $f$ such that $\vect{z}^* = \underset{\vect{z} \in \mathcal{Z}}{\text{argmin }} f(\vect{z})$ corresponds to the desired experimental result. 
	The optimization task can be approached with a closed-loop strategy, which iteratively evaluates a set of options $\vect{z}_j$ and records associated responses, $f_j = f(\vect{z}_j)$, to gradually collect a set of observations, $\mathcal{D}_n = \{ \vect{z}_j, f_j \}_{j=1}^n$, as feedback to the experiment planning strategy. 
	
	\begin{figure}[htb]
		\centering
		\includegraphics[width=1.0\columnwidth]{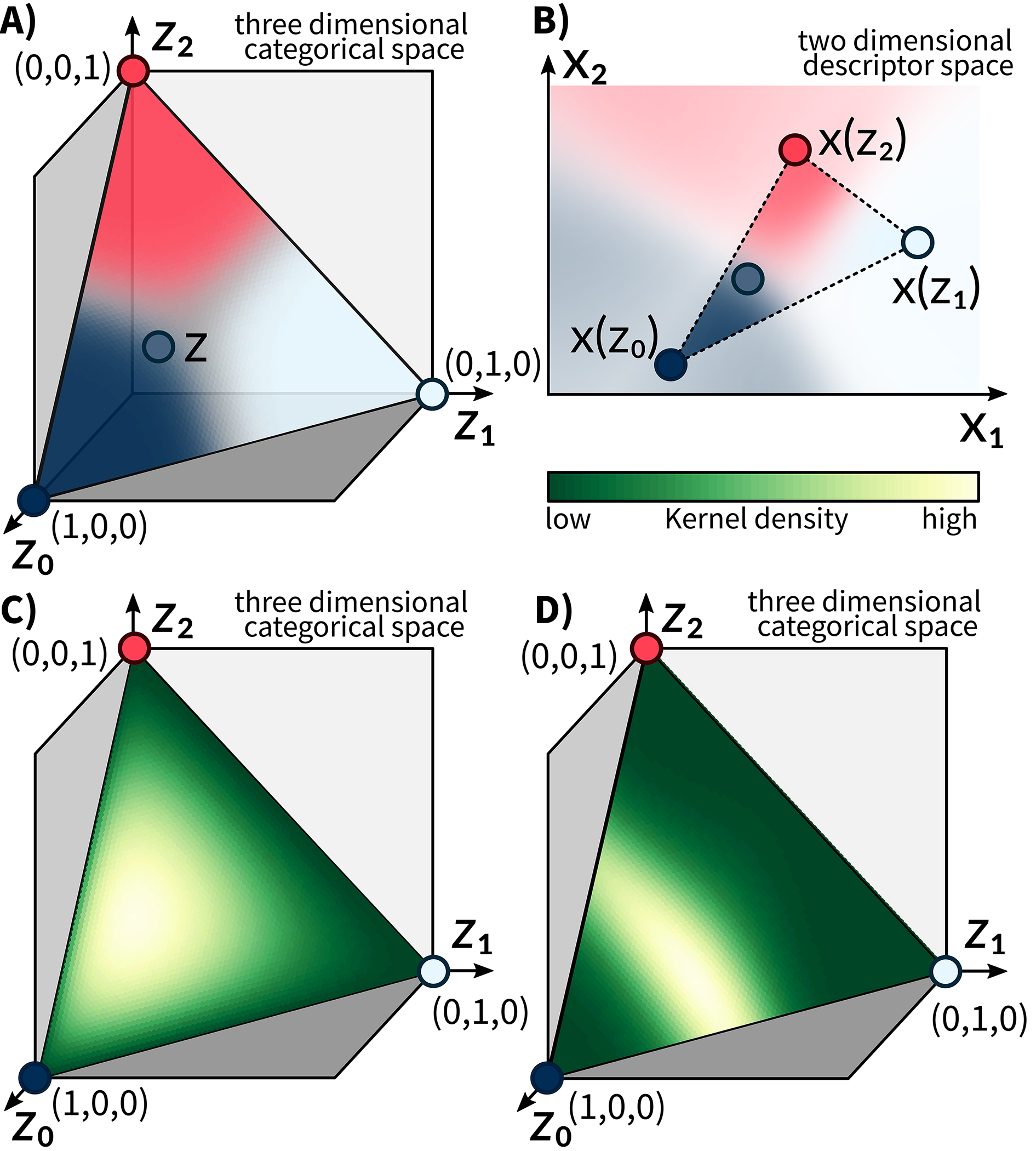}
		\caption{Illustrations of the na\"ive and the static \gryffin strategies for the (descriptor-guided) optimization of categorical variables. (A) Illustration of a categorical variable with three options represented on a simplex. Color contours indicate the affiliation of any given point on the simplex to one of the categorical options at the corners. (B) Representation of a continuous descriptor space, where descriptors are associated with the categorical options shown in panel A. Color contours indicate the affiliation with individual options of the categorical variable. (C) Illustration of a generic kernel density on the simplex modeled with a concrete distribution. (D) Illustration of a descriptor-guided transformation of the kernel density shown in panel C based on the descriptors shown in panel B.}
		\label{fig:space_illustration}
	\end{figure}

	The optimization of categorical parameters poses additional challenges compared to the optimization of continuous or discrete variables due to the lack of a natural ordering between individual parameter values, which is illustrated in the supplementary information (see Sec.~\ref{supp_sec:benchmark_functions}). Regardless of the optimization strategy, the confidence of having identified the best performing candidate in the search space increases with the number of evaluated candidates. In the best case, only one evaluation is required, while in the worst case all candidates in the search space need to be evaluated. Yet, the choice of the optimization strategy modulates the chance of having identified the best performing candidate after a certain number of evaluations and consequently the average fraction of the candidate space that needs to be evaluated to identify the most desired one.

	Straightforward search strategies rely on exhaustive random\cite{Bergstra:2012, Baba:1981, Matyas:1965} or systematic\cite{Anderson:2016, Box:2005, Fisher:1937} evaluations of all candidates without leveraging any feedback from collected responses to refine the search policy. 
	In the absence of accurate prior expectations on the performance of individual candidates, both random and systematic search strategies require the evaluation of \unit[50]{\%} of all candidates, on average, to identify the best performing candidate and are therefore only applicable to relatively small search spaces. 
	Yet, exhaustive strategies are massively parallelizable and thus well suited for high-throughput experimentation. 
	Genetic algorithms and evolutionary strategies\cite{Holland:1989, Koza:1992, Srivinas:1994} extend the idea of a random exploration of the search space. 
	While they typically start with random selection of candidate solutions, they then condition exploration policies based on a population of candidate solutions whose fitness has already been evaluated.
	Poorly performing candidates in the population are dropped, while new ones are selected based on local perturbations of the best performing candidates. 
	Thus, the set of new candidate solutions to be evaluated is constantly updated based on a local, rather than global, search strategy.\cite{Mirjalili:2020}

	Bayesian optimization\cite{Mockus:1978_bopt, Kushner:1964_bopt} has recently gained increasing attention as a competitive global optimization strategy across various fields, \cite{Shahriari:2015, Jones:2001_bopt,Pyzer-Knapp2018} including automatic machine learning,\cite{Feurer:2015, Thornton:2013, Swersky:2013} and experimental design.\cite{Von:2019, Foster:2019, Vanlier:2012}
	The common framework of Bayesian optimization strategies follows two basic steps: (i) the construction of a surrogate to the unknown response surface from a probabilistic model based on collected measurements, and (ii) the selection of new candidates with an acquisition function that balances the expected performance of each candidate and the uncertainty of this estimate. 
	Various machine learning models have been suggested to construct the surrogate, including Gaussian processes,\cite{Rasmussen:2003} random forests\cite{Breiman:2001}, and Bayesian neural networks,\cite{Snoek:2015} and different acquisitions functions, such as probability of improvement,\cite{Kushner:1964_bopt} expected improvement,\cite{Jones:1998_bopt} upper (lower) confidence bound\cite{Srinivas:2012_bopt}, predictive entropy,\cite{Hernandez:2014_bopt} and kernel density based formulations,\cite{Hase:2018_phoenics} are commonly employed.

	Extensions of Bayesian optimization frameworks to categorical parameter domains are under active development. 
	One approach consists in the representation of categorical parameters as one-hot encoded vectors.\cite{Golovin:2017_bopt, Gpyopt:2016, Snoek:2012_bopt} 
	This representation expresses the $j$-th option of a categorical variable $\vect{z}$ with $n$ different options, $\vect{z} = \{ z_1, \ldots, z_n\}$, as an $n$-dimensional vector with elements $z_i = \delta_{ij}$ for $1 \leq i \leq n$, which can be interpreted as the corners of an $n$-dimensional simplex, $\vect{z} \in \Delta^{n-1} = \{ \vect{z} \in \mathbb{R}^n | z_i \in [0, 1] \text{ and } \sum_{i=1}^n z_i = 1\}$ (see Fig.~\ref{fig:space_illustration}a).
	Standard Bayesian optimization strategies for continuous parameter domains can be deployed on these one-hot encoded categorical variables, such that even optimizations of mixed continuous-categorical domains are possible. 
	However, the most promising choices for future evaluations are determined by projecting promising candidates from the continuous space to the one-hot boundaries.
	This strategy presents two limitations. 
	First, redundancies in the projection arise from the fact that the continuous optimization space contains an additional degree of freedom compared to the categorical domain. 
	These redundancies can be reduced by imposing constraints on the acquisition function. 
	For example, the acquisition function can be modified such that covariances are computed after the projection operation.\cite{Garrido:2019_bopt, Hernandez:2017_bopt} 
	This modification results in a stepwise acquisition function from which choices for future evaluations can be suggested directly. 
	However, stepwise functions are generally more challenging to optimize than smooth ones. 
	A second limitation is due to the fact that one-hot encoding imposes an equal measure of covariance between all choices of the categorical variables. 
	As a consequence, all categorical options are considered equally similar to each other, and one cannot account for imbalanced similarities. 
	Mixed categorical-continuous Bayesian optimization displays similar challenges and is the subject of current research too.\cite{Daxberger:2020,Ru:2020}
	
	Because of the importance of categorical variables in chemistry and materials science, recent advances in Bayesian optimization algorithms have sought to overcome the aforementioned issues to find application in reaction optimization\cite{Shields:2021} and materials design\cite{Zhang:2020}. Shields \textit{et al.}\cite{Shields:2021} encoded categorical variables as continuous descriptor vectors and pruned highly-correlated descriptors. Zhang \textit{et al.}\cite{Zhang:2020b,Zhang:2020} have proposed a latent-variable Gaussian process approach that maps categorical options to a low-dimensional numerical latent space. In both cases, these continuous representations reflected chemico-physical properties of their respective categorical options (e.g., solvent, catalyst, metal ion, etc.), such that a measure of similarity between options could be established. These approaches were shown to be superior to using one-hot encoded vectors to describe the categorical options.

	\section{Formulating \gryffin}

	Building upon previous work (see Sec.~\ref{sec:background}), we base the formulation of \gryffin on a one-hot encoding of categorical variables. 
	Yet, rather than constructing the surrogate on the continuous space spanned by the one-hot encoded categorical choices, where each dimension is bounded by $[0, 1]$, \gryffin aims to support the surrogate on the simplex to avoid projection redundancies. 
	To this end, we extend the recently reported \phoenics approach\cite{Hase:2018_phoenics} from continuous to categorical domains. 
	
	\phoenics uses kernel regression, based on the Nadaraya-Watson estimator\cite{Nadaraya:1964,Watson:1964}, to build a surrogate model of the objective function. The kernel density estimates used by this model are inferred by a Bayesian neural network (BNN) with an autoencoder-like architecture. The use of this BNN allows us to obtain complex multivariate density estimates that are regularized by the prior distribution of the kernels’ precision. Because the BNN is used to estimate kernel densities only, evaluation of the objective function does not require additional BNN evaluations. \phoenics’s acquisition function is constructed by adding a single uniform kernel, weighted by a user-defined $\lambda$ parameter, to the surrogate model. This uniform kernel can be interpreted as the lack of knowledge present in regions of parameter space with low density of observations, following the assumption that the global optimum may be located anywhere in the search domain with equal probability. The $\lambda$ parameter offers an intuitive way to bias the behavior of the optimizer towards exploration or exploitation. It also provides a simple yet effective means of enabling batch optimization, given that acquisitions with multiple different $\lambda$ values (each with a different exploration-exploitation trade-off) can be readily obtained. Overall, this framework results in a Bayesian optimization approach that scales linearly with the number of observations and the dimensionality of the search space, and that naturally supports optimization in parallel.\cite{Hase:2018_phoenics}
	
	Similar to \phoenics, in \gryffin we model categorical parameters as random variables and construct the surrogate from their reweighted kernel density estimates as inferred by a BNN.
	Beyond the implementation of kernel density-based Bayesian optimization on categorical domains, we further demonstrate how physical and chemical domain knowledge can be used to transform the surrogate to accelerate the search, and how this bias can be refined during the optimization to gain scientific insight.

	\subsection{Categorical optimization with na\"ive \gryffin}

		Na\"ive \gryffin constructs kernel densities by extending the one-hot encoding of categorical options to the entire simplex, i.e.~we consider $\vect{z} \in \Delta^{n-1}$. The largest entry of any given point $\vect{z}$ can be used to associate this point to a realizable option $\vect{\zeta}$ (see Fig.~\ref{fig:space_illustration}a). Various probability distributions with support on the simplex have been introduced in the past. The Dirichlet distribution, for example, constitutes the conjugate prior to the categorical distribution.\cite{Ng:2011} Another example is the logistic normal distribution which ensures that the logit of generated samples follow a standard normal distribution.\cite{Atchison:1980} While both of these distributions are widely used, their deployment in a computational graph is numerically involved due to demanding inference and sampling steps. Directed probabilistic models can be implemented at low computational cost if stochastic nodes of such graphs can be reparameterized into deterministic functions of their parameters and stationary noise distributions.\cite{Kingma:2013} Such reparameterizations, however, are unknown for the Dirichlet and the logistic normal distribution. 

		The recently introduced concrete distribution\cite{Maddison:2016} (simultaneously introduced as Gumbel-Softmax),\cite{Jang:2016} illustrated in Fig.~\ref{fig:space_illustration}c, overcomes this obstacle. This distribution is supported on the simplex and parameterized by a set of deterministic variables with noise generated from stationary sources. As such, the concrete distribution is amenable to automatic differentiation frameworks for accelerated sampling and inference. In addition, the concrete distribution contains a temperature parameter, $\tau$, which can be tuned to smoothly interpolate between the discrete categorical distribution and a uniform distribution on the simplex. As such, this temperature parameter controls the localization of constructed kernel densities towards the corners of the simplex.

		Na\"ive \gryffin estimates kernel densities from concrete distributions and conditions the parameters of the concrete distribution on the sampled candidates, as introduced by the \phoenics framework.\cite{Hase:2018_phoenics} The temperature parameter is modified based on the number $n$ of collected observations, $\tau \sim n^{-1}$, such that the priors gradually transition from a uniform distribution to a continuous approximation of the categorical distribution. Options for future evaluation are determined via the acquisition function of \phoenics, which compares the constructed kernel densities to the uniform distribution on the simplex. Using a sampling parameter $\lambda$ to reweight the uniform distribution, this acquisition function can favor exploration or exploitation explicitly, and natively supports batch optimization. Details of this procedure are provided in the supplementary information (see Sec.~\ref{supp_sec:naive_gryffin_derivation}).

		Approximating the computation of kernel densities can further reduce the overall computational cost of \gryffin by a significant amount. The approximation is based on the idea that the low density regions of the kernel densities indicate a lack of information. A precise estimate of the kernel densities in these regions might therefore not be required. We find that an approximate estimate of the kernel densities in low density regions can significantly reduce \gryffin's computational cost without degrading optimization performance. Details on this approximation are provided in the supplementary information (see Sec.~\ref{supp_sec:boosting})

	\subsection{Descriptor-guided searches with static \gryffin}

		The na\"ive \gryffin approach imposes an equal measure of covariance between individual options of categorical variables, which is undesired in cases where a notion of similarity can be established between any two options. Especially in the context of scientific discovery, similarities between the options of categorical variables can be defined, for example, \emph{via} physicochemical descriptors for small molecules or material candidates. We extend the na\"ive approach by assuming that the metric to measure similarity between any two options is based on the Euclidean distance between real-valued $d$-dimensional descriptor vectors, $\vect{x} \in \mathbb{R}^d$, which are uniquely associated with individual categorical options (see Fig.~\ref{fig:space_illustration}a,b). 

		While the descriptors are embedded in a continuous space, their arrangement in this space is unknown for a generic optimization task and only selected points in the descriptor space can be associated with realizable categorical options. These limitations present major obstacles to optimization strategies that operate directly on the descriptor space. Instead, we propose to leverage the na\"ive \gryffin framework but redefine the metric on the simplex based on the provided descriptors. Following this strategy, the length of an infinitesimal line element on the simplex is conditioned not only on the corresponding infinitesimal change of location on the simplex, but also on the infinitesimal change of the associated descriptors. Imposing a linear mapping between points on the simplex $\vect{z}$ and the descriptor space $\vect{x}$, we can compute the length of a finite line element, $\Delta s$, following the redefined metric to be
		\begin{align}
			\Delta s^2 = \sum\limits_{m = 1}^{\text{\#descs}} \sum\limits_{i,j=1}^\text{\#opts} \left( x_m^i - \sum\limits_{k=1}^\text{\#opts} z_k x_m^k \right)^2,
		\end{align}
		which is derived in detail in the supplementary information (see Sec.~\ref{supp_sec:static_gryffin_derivation}). Kernel densities generated by the na\"ive approach can be transformed following this descriptor-based definition of distances on the simplex to reflect the similarity between individual options as illustrated in Fig.~\ref{fig:space_illustration}a,b. As a consequence, the evaluation of one option of the categorical variable will be more informative with respect to the expected performance of other, similar options. Further implications of imposing a descriptor-guided metric on the simplex are illustrated in the supplementary information (see Sec.~\ref{supp_sec:static_gryffin_derivation}).

		We refer to the descriptor-guided categorical optimization as static \gryffin, as provided descriptors are used without further transformations.
		The benefits of static \gryffin over na\"ive \gryffin with respect to an accelerated search will depend on the provided descriptors. 
		More informative descriptors are expected to guide the algorithm toward well-performing options more efficiently than less informative descriptors. 
		This effect is investigated empirically in Sec.~\ref{sec:results}.

	\subsection{Descriptor refinement with dynamic \gryffin}

		The dynamic formulation of \gryffin aims to alleviate the expected sensitivity of the performance of static \gryffin on the choice of provided descriptors by transforming them during the optimization. Specifically, dynamic \gryffin infers a transformation, $T$, which constructs a new set of descriptors, $\vect{x}^\prime$, from the provided descriptors, $\vect{x}$, based on the feedback collected from evaluated options. The transformation $T$ can be constructed to target two major goals: (i) the generation of more informative descriptors, $\vect{x}^\prime$, which help to navigate the candidate space more efficiently, and (ii) the interpretable identification of relevant domain knowledge to inspire design choices and scientific insights, as we will demonstrate in Sec.~\ref{sec:experiments}. In addition to these two goals, the transformation $T$ is required to be robust with respect to overfitting due to the low data scenarios which are commonly encountered in autonomous workflows. 

		In an attempt to balance flexibility, robustness and interpretability, we suggest constructing this transformation $T$ from a learnable combination of the provided descriptors
		\begin{align}
		\label{eq:descriptor_loss}
			\vect{x}^\prime = \text{softsign} \left( \vect{W} \cdot \vect{x} + \vect{b} \right),\quad \text{softsign}(x) = \frac{x}{1 + |x|},
		\end{align}
		where $\vect{W}$ and $\vect{b}$ are the learnable parameters inferred from the feedback collected in previous evaluations. The class of transformations described by Eq.~\ref{eq:descriptor_loss} includes slightly non-linear translations and rotations of the provided descriptors. While more complex transformations accounting for higher-order interactions between individual descriptors could potentially yield even more informative descriptors, a slightly non-linear transformation is inherently robust to overfitting,\cite{Geman:1992} and is more amenable to intuitive interpretation than more complex models.\cite{Molnar:2019} We will demonstrate empirically in Secs.~\ref{sec:results} and \ref{sec:experiments} that this class of transformations is well suited for a variety of categorical optimization tasks. 

		Following a stochastic gradient optimization, the parameters $\vect{W}$ and $\vect{b}$ in Eq.~\ref{eq:descriptor_loss} are adjusted to (i) increase the correlation between the newly generated descriptors $\vect{x}^\prime$ and the associated measurements, (ii) reduce correlations between newly generated descriptors, and (iii) remove redundant descriptors with poor correlations with the measurements or high correlations with other newly generated descriptors. These three goals are modeled as penalties which are to be minimized at training time (see supplementary information Sec.~\ref{supp_sec:dynamic_gryffin_derivation} for details).

	\section{Synthetic benchmarks}
\label{sec:results}

	We empirically assess the performance of all proposed variants of \gryffin on a set of synthetic benchmark surfaces, which are detailed in the supplementary information (see Sec.~\ref{supp_sec:benchmark_functions}). Four of the surfaces constitute categorized adaptations of established functions commonly used to benchmark global and local optimization strategies on continuous parameter domains. In addition, we include three partially and fully randomized surfaces with responses sampled from stationary probability distributions. While the ordering of the categorical options is arbitrary, we introduce a reference ordering to illustrate the surfaces (see supplementary information, Sec.~\ref{supp_sec:benchmark_functions}, for details). Unless noted otherwise, descriptors for the categorical options are generated such that they encode the reference ordering. Implementations of all benchmark surfaces are made available on GitHub.\cite{github_repo} 

	\gryffin is compared to a set of qualitatively different optimization strategies that are implemented in publicly available libraries: genetic optimization available through \emph{PyEvolve},\cite{Perone:2009, Koza:1992, Holland:1989} Bayesian optimization with random forests as implemented in \emph{SMAC},\cite{smac:2017, Hutter:2012, Hutter:2011} Bayesian optimization with Gaussian processes \emph{via} \emph{GPyOpt},\cite{Gpyopt:2016, Gonzalez:2016_glasses, Gonzalez:2016_batch, Gonzalez:2015} and Bayesian optimization with tree-structured Parzen windows introduced in \emph{Hyperopt}.\cite{Bergstra:2013, Bergstra:2011} In addition, we run random explorations of the candidate space as a baseline. We compare the performance of the different formulations of \gryffin to the other optimization strategies on all benchmark surfaces, probe the influence of the number of descriptors, study the scaling of \gryffin with the number of options per categorical variable and the number of categorical variables, and investigate the benefits of dynamic \gryffin over static \gryffin. For all comparisons, we measure the fraction of the candidate space that a given optimization strategy explored to locate the best candidate. Unless noted otherwise, all comparisons are averaged over $200$ independent executions of each strategy. 

	\begin{figure}[htb]
		\centering
		\includegraphics[width=1.0\columnwidth]{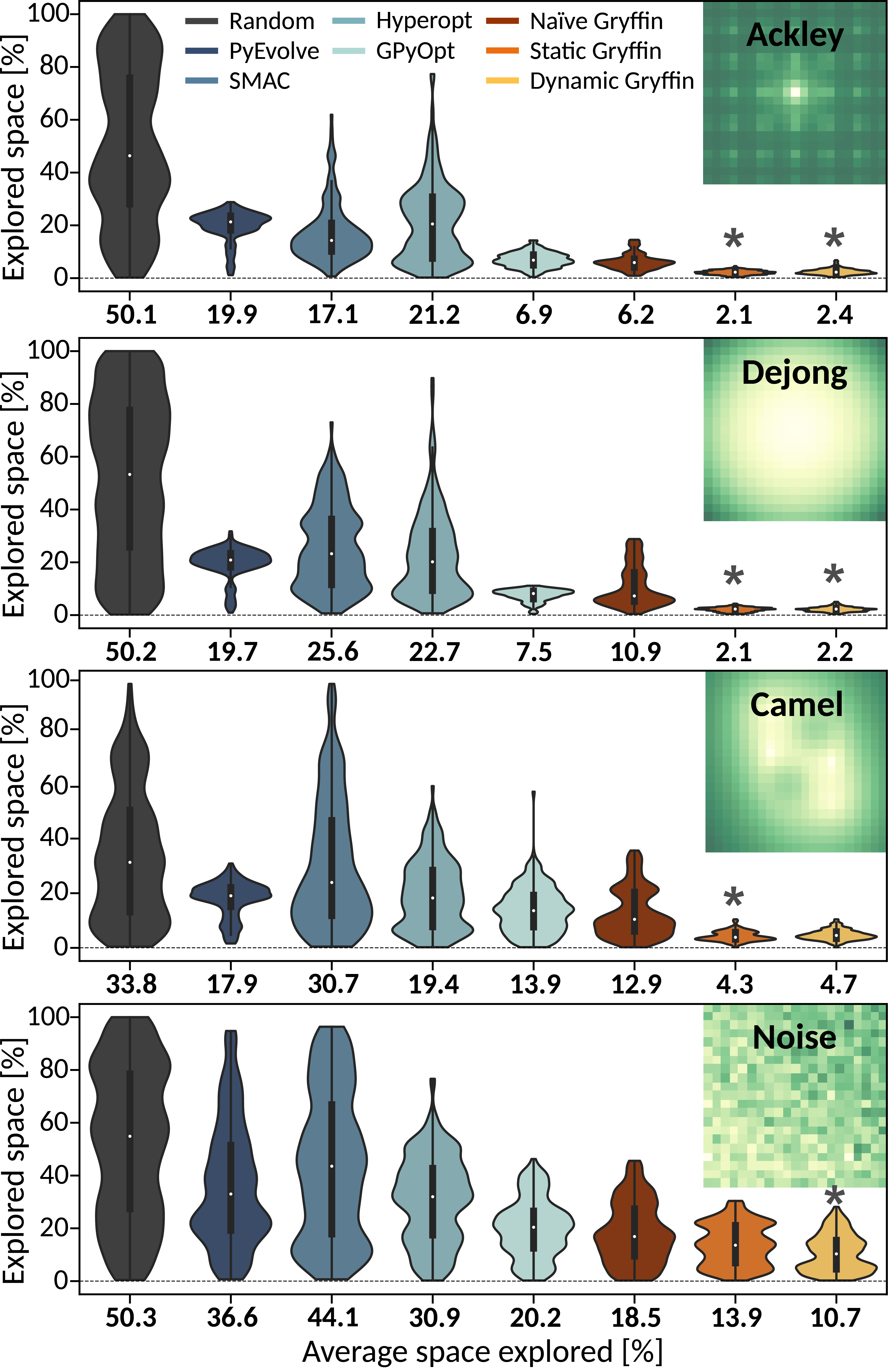}
		\caption{Performance of various optimization strategies on selected synthetic surfaces. Individual panels indicate the fraction of the candidate space each algorithm explored before finding the global minimum of the surfaces illustrated on the top right (low values are shown in yellow, high values in green), averaged over $200$ independent executions. Best performing algorithms are indicated by a star. Color codes for the optimization strategies are shown in the top panel and apply to all panels in this figure.}
		\label{fig:analytic_benchmarks}
	\end{figure}

	\subsection{Optimization performance}
	\label{sec:analytic_benchmarks}

		In a first test, we compare the optimization strategies on two-dimensional formulations of the synthetic benchmark surfaces with $21$ options per dimension, as illustrated in Fig.~\ref{fig:analytic_benchmarks}. 
		The \emph{Dejong} surface generalizes the convex parabola from continuous to categorical spaces, such that we consider it as \emph{pseudo-convex}. 
		In contrast, the \emph{Ackley} surface is generalized from the Ackley path function which is non-convex on the parameter domain. 
		The \emph{Camel} surface presents a degenerate global optimum as there are two different combinations of options which yield the same optimal response. Finally, in the \emph{Noise} surface, the response increases linearly with the index of the option along each dimension in the reference ordering (as in \textit{Slope}, see \ref{supp_sec:benchmark_functions}), but uniform noise is added to response to yield a substantial local variance. 
		The fractions of the search spaces that the optimizers explored to locate their optima, averaged over $200$ independent executions, are illustrated in Fig.~\ref{fig:analytic_benchmarks}. 
		Full optimization traces are reported in the supplementary information (see Sec.~\ref{supp_sec:analytic_benchmarks_traces}).

		We observe that random exploration requires the evaluation of approximately half the space for the surfaces with well-defined global optima, and about a third of the space for the \emph{Camel} function with degenerate optima. 
		We find that the performances of \emph{PyEvolve}, \emph{SMAC} and \emph{Hyperopt} are roughly comparable across the different surfaces, although \emph{PyEvolve} tends to outperform \emph{SMAC} and \emph{Hyperopt} on the noiseless surfaces. 
		\emph{GPyOpt} generally locates global optima faster than the other strategies, but is slightly slower than na\"ive \gryffin on the non-convex surfaces. 
		The faster optimization of convex surfaces by Gaussian process-based Bayesian optimization, compared to kernel density-based approaches, has already been observed and discussed for continuous domains.\cite{Hase:2018_phoenics} 
		Notably, the static and dynamic formulations of \gryffin can significantly outperform the other optimization strategies, with reductions of the explored space by several factors. 
		This observation confirms that providing real-valued descriptors can substantially accelerate the search. 
		We also observe similar performances of the static and dynamic formulations of \gryffin for the deterministic surfaces (\emph{Ackley}, \emph{Dejong}, \emph{Camel}), while dynamic \gryffin optimizes the noisy surface at a faster rate. 
		This observation suggests that dynamic \gryffin is indeed capable of learning a more informative set of descriptors.

	\subsection{Scaling to more options and higher dimensions}

		We further study the performance of \gryffin on larger candidate spaces with (i) more categorical variables, and (ii) more options per categorical variable. 
		Increasing the number of variables or the number of options per variable generally increases the number of candidates in the space and is thus expected to require more candidate evaluations overall before the best candidate is identified. 
		Results obtained for these benchmarks are detailed in the supplementary information (see Secs.~\ref{supp_sec:analytic_benchmarks_dimensions} and \ref{supp_sec:analytic_benchmarks_options}). 
		The benchmarks suggest that \gryffin indeed uses more candidate evaluations to locate the global optimum with an increasing volume of the search space, consistently across all benchmark surfaces. 
		However, although the number of evaluations increases, the fraction of the explored space generally decreases. 
		More specifically, across the different surfaces, we identify a polynomial decay of the explored space with an increasing number of options per variable, with decay exponents ranging from $-1.0$ to $-1.25$, and an exponential decay for an increase in the number of parameters with decay coefficients ranging from $-1.6$ to $-2.0$ (more details in the supplementary information, Secs.~\ref{supp_sec:analytic_benchmarks_dimensions} and \ref{supp_sec:analytic_benchmarks_options}) 
		Based on this observation, we conclude that \gryffin may show an onset of the \emph{curse of dimensionality}\cite{Bellman:2015} only for a relatively large number of dimensions and thus constitutes an optimization strategy that can efficiently navigate large categorical spaces.

	\subsection{Data-driven refinement of descriptors}
	\label{sec:learning_descriptors}

\begin{figure*}[!htb]
			\centering
			\includegraphics[width=1.0\textwidth]{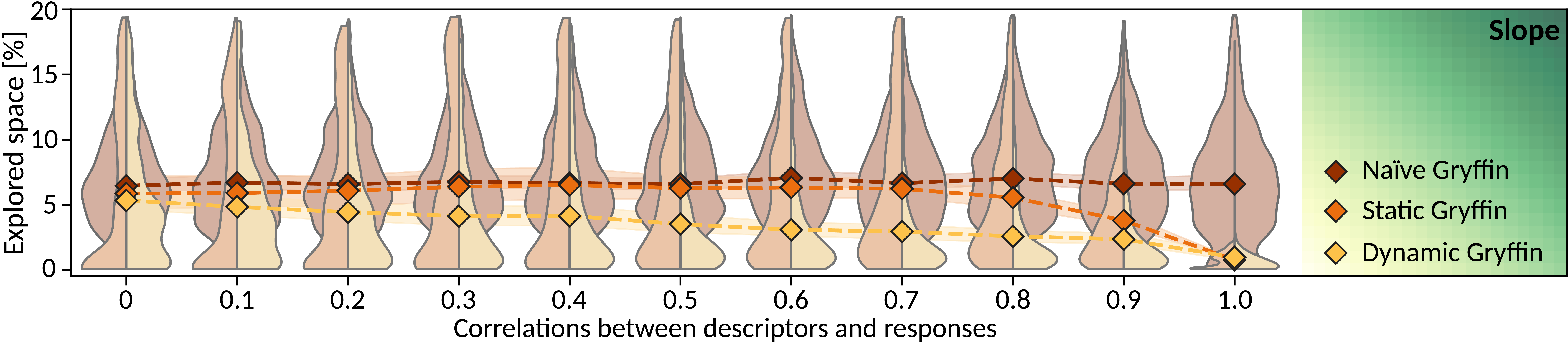}
			\caption{Behavior of na\"ive, static and dynamic \gryffin on the \textit{Slope} surface for different descriptors. Descriptors have been generated randomly with a targeted correlation between the descriptors and the response of the associated options. The correlation between descriptors and response is shown on the x-axis. The y-axis represents the fraction of the search space explored before the identification of the optimum. Kernel density estimates show the distribution of explored space fractions across $200$ repeated simulations, while square markers represent the mean. The performance of na\"{i}ve \gryffin is independent of the descriptor correlation, as expected. Static and dynamic \gryffin always showed equal or superior performance to the na\"{i}ve formulation. Static \gryffin outperformed the na\"{i}ve formulation for  correlations equal to or above $0.8$. Dynamic \gryffin showed superior performance to the na\"{i}ve formulation for any correlation equal to or above $0.1$, suggesting that  dynamic \gryffin is able to take advantage of even poorly informative descriptors to improve the optimization performance.}
			\label{fig:descriptor_learning}
		\end{figure*}

		The effectiveness of transforming provided descriptors to accelerate the search for the best candidate is studied in detail on the \emph{Slope} surface with $51$ options per dimension, resulting in $2,601$ different candidates (see Fig.~\ref{fig:descriptor_learning}). For this benchmark, we randomly assign descriptors to each of the categorical options at a desired targeted correlation between the descriptors and the responses of the associated options. With a decreasing correlation, the local variance increases, which results in a less structured space that is more challenging to navigate. We therefore generally expect a performance degradation for both static and dynamic \gryffin with decreasing correlation.

		Fig.~\ref{fig:descriptor_learning} illustrates the fractions of the candidate space explored by static and dynamic \gryffin for different targeted correlations between the supplied descriptors and the responses. 
		For comparison, we also report the performance of the na\"ive formulation of \gryffin, which is independent of the supplied descriptors. 
		We observe a significant increase in the fraction of the explored space with decreasing correlation for both static and dynamic \gryffin. 
		Although both methods require more candidate evaluations with less informative descriptors, their performance never degrades beyond the performance of the na\"ive formulation, indicating that even entirely uninformative descriptors do not delay the search for the best candidate compared to descriptor-less scenarios. Note that, for the purpose of redefining the metric on the simplex, negatively correlated descriptors are as informative as positively correlated ones, both for static and dynamic \gryffin.

		We further find that static \gryffin can benefit from descriptors and significantly outperform the na\"ive approach if the Pearson correlation coefficient between descriptors and responses is at least $0.8$. Below this value, the average performance of static and na\"ive \gryffin is comparable, although the variance on the performance is higher for the static formulation. Similar to static \gryffin, learning a more informative set of descriptors with dynamic \gryffin accelerates the search more if the correlation between the descriptors and the responses is high. However, the dynamic formulation is generally at least as fast as the static formulation and can successfully leverage descriptors to outperform descriptor-less searches even at correlations as low as $0.1$. We thus confirm that the descriptor transformation introduced in Eq.~\ref{eq:descriptor_loss} is sufficiently robust to be applied to low-data tasks and conclude that deploying dynamic \gryffin can be beneficial for some descriptor guided optimization tasks without delaying the optimization compared to static \gryffin. 

	\section{Applicability of \gryffin to chemistry and materials science}
\label{sec:experiments}

	Following the empirical benchmarks of \gryffin, we now demonstrate its applicability and practical relevance to a set of optimization tasks across materials science and chemistry. Specifically, we target the discovery of non-fullerene acceptors for organic solar cells, the design of organic-inorganic perovskites for light harvesting, and the selection of phosphine ligands and optimization of process conditions for Suzuki-Miyaura coupling reactions. 

	Obtaining statistically significant performance comparisons at a sufficient level of confidence requires the repeated execution of optimization runs to average out the influences of initial conditions and probabilistic elements of the optimization strategies. As repetitive executions of optimization runs on these applications are highly resource demanding, we construct these optimization tasks from recently reported datasets: the applications on the discovery of non-fullerene acceptors and perovskites are based on lookup tables, and the optimization of Suzuki reactions are facilitated via a probabilistic model trained on experimental data (\emph{virtual robot}) to emulate experimental uncertainties in addition to the average response. Virtual robots have recently been introduced to benchmark algorithms for autonomous experimentation.\cite{Hase:2018_chimera,Hase:2020_olympus} 

	The selection of physicochemical descriptors associated with each option of the categorical variables were based on chemical intuition as well as their accessibility. The most informative set of descriptors for a specific task might be \emph{a priori} unknown. However, a domain expert might have an intuition for which descriptors might be informative for the task at hand. In addition, given that descriptors need to be provided for all candidate solutions in the search space, it is convenient for these to be easily accessible. As such, in the following sections, descriptors are selected based on their expected suitability to each task as well as their availability.

	\subsection{Discovery of non-fullerene acceptor candidates for organic photovoltaics}
	\label{sec:non_fullerenes}

		Small organic molecules currently constitute the highest performing acceptor materials for organic solar cells.\cite{Yan:2018_opv, Hou:2018_opv} The large number of degrees of freedom when designing such non-fullerene acceptors, arising from complex aromatic moieties, allows to fine tune their relevant electronic properties, for example the optical gap and the energy level alignment between the donor and acceptor materials. While this large design space provides the required flexibility to fine-tune desired molecular properties, it is also challenging to navigate and constitutes a major obstacle to the discovery of novel candidate molecules.

		\begin{figure*}[htb]
			\centering
			\includegraphics[width = 1.0\textwidth]{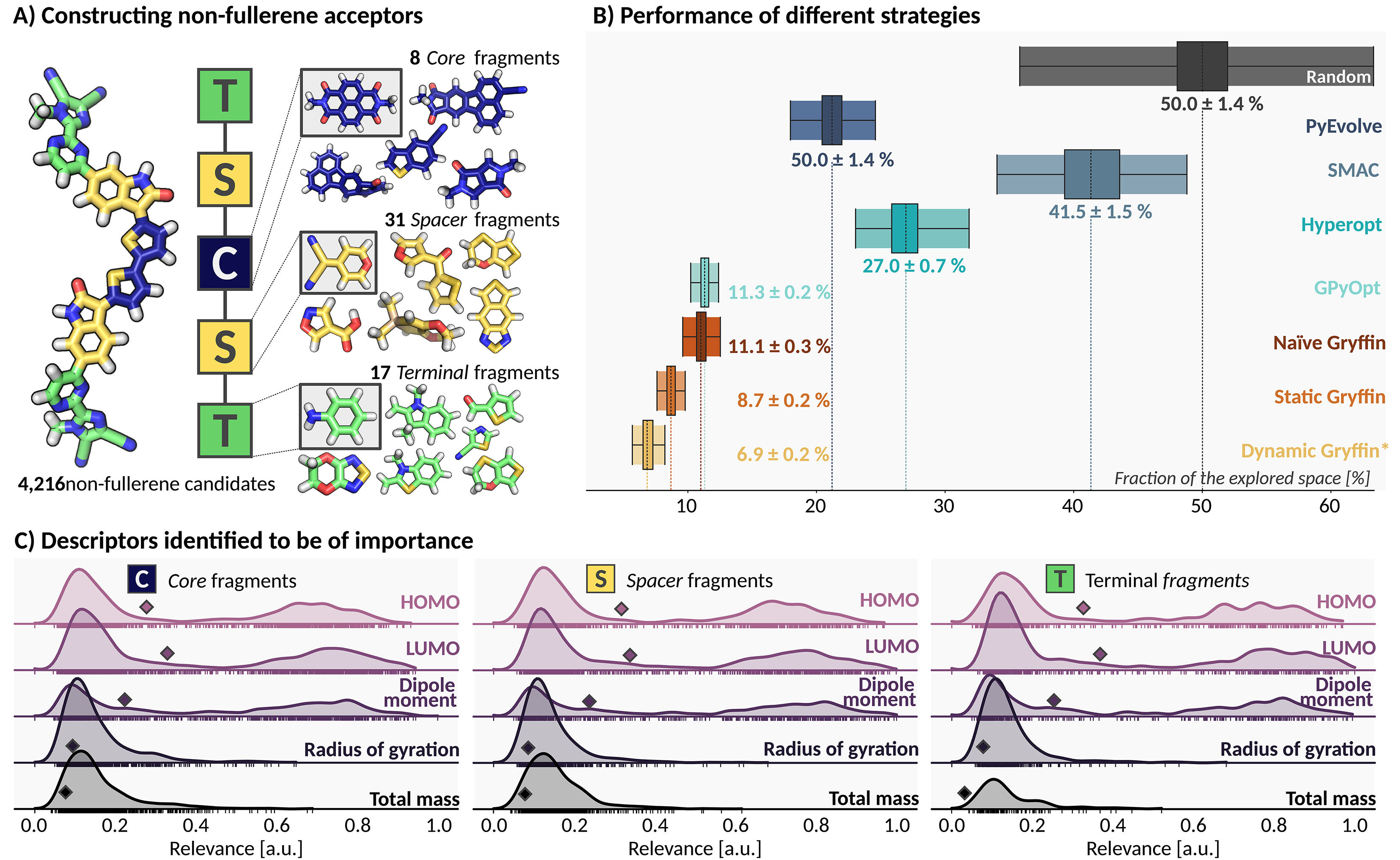}
			\caption{Performance of \gryffin on the task of identifying non-fullerene acceptors for maximized power conversion efficiencies. (A) Non-fullerene acceptor candidates are constructed from a set of molecular fragments (core, spacer and terminal) which are symmetrically arranged to span a library of 4,216 different candidate molecules. (B) Fraction of the candidate library to be explored by each of the studied optimization strategies to identify the best performing acceptor candidate. (C) Most informative descriptors to guide the search of \emph{dynamic} \gryffin. The relevance metric is computed as the normalized weights that \emph{dynamic} \gryffin learnt to associate to each descriptor. The distributions shown are over the $200$ repeated optimizations performed. Diamonds indicate the average relevance of each descriptor.}
			\label{fig:app_non_fullenere_results}
		\end{figure*}

		We demonstrate the applicability of \gryffin for the discovery of non-fullerene acceptors on a candidate space of 4,216 different small organic molecules, which form a subset of a recently reported comprehensive study.\cite{Lopez:2017} Acceptor candidates in this library are constructed from a set of molecular fragments that are separated into three fragment pools (see Fig.~\ref{fig:app_non_fullenere_results}a). Each candidate is composed of one core fragment $C$ (8 options), two spacer fragments $S$ (31 options), and two terminal fragments $T$ (17 options) following a symmetric design. Details on the library of candidate fragments are reported in the supplementary information (see Sec.~\ref{supp_sec:non_fullerene_results}). The performance of each acceptor candidate is quantified based on the power conversion efficiency (PCE), computed from calibrated DFT frontier molecular orbital energies.\cite{Lopez:2017} The optimization task aims to identify molecules with the highest possible PCE.

		We guide static and dynamic \gryffin with a set of electronic and geometric descriptors for each of the fragments: the HOMO and LUMO energy levels, the dipole moment, the radius of gyration, and the molecular weight. Electronic properties were computed at the B3LYP/Def2SVP level of theory on a SuperFineGrid using Gaussian,\cite{gaussian:2016} and the radius of gyration was computed for the ground state conformations of the fragments.
		The correlations of the descriptors with the PCE of the resulting non-fullerene acceptor are generally low, with the highest encountered Pearson correlation coefficients reaching values of about $0.2$ (see supplementary information, Sec.~\ref{supp_sec:non_fullerene_results}, for details). 
		In fact, the identification of improved descriptors for the accurate prediction of PCE in organic solar cells is an active field of research.\cite{Sahu:2018_opv, Venkatraman:2015_opv, Scharber:2006_opv}

		Fig.~\ref{fig:app_non_fullenere_results}b illustrates the fraction of the candidate library (averaged over $200$ independent executions) that each optimization strategy had to explore before identifying the combination of fragments yielding the highest PCE. Full optimization traces for all optimization strategies are reported in the supplementary information (see Sec.~\ref{supp_sec:non_fullerene_results}). 
		In agreement with the synthetic tests (see Sec.~\ref{sec:results}), we find that \emph{PyEvolve} explores a smaller fraction (\unit[21]{\%}) of the space than \emph{Hyperopt} (\unit[27]{\%}) or \emph{SMAC} (\unit[41]{\%}) before identifying the best candidate molecule. 
		The performance of na\"ive \gryffin, exploring about \unit[11]{\%}, is comparable to \emph{GPyOpt} and thus significantly faster than the other benchmark strategies. However, the physical descriptors supplied for each of the fragments enable static \gryffin to find the best acceptor candidate after exploring only \unit[8.7]{\%} of the candidate space (\unit[$\sim 22$]{\%} reduction of the required acceptor evaluations compared to na\"ive search). 
		Dynamic \gryffin can refine the supplied descriptors to find the best candidate while exploring only \unit[6.9]{\%} of the library (\unit[$\sim 38$]{\%} reduction over na\"ive search).
		This improvement of dynamic \gryffin over static \gryffin confirms that the supplied descriptors can be transformed into a more informative set to accelerate the search.

		Fig.~\ref{fig:app_non_fullenere_results}c illustrates the importance of individual descriptors in guiding the search, as determined by \emph{dynamic} \gryffin. Specifically, we plot the relative contributions of individual descriptors to the set of the transformed descriptors that were used when the best performing candidate was identified. We observe that the descriptor-based search emphasizes the relevance of electronic descriptors over geometric descriptors consistently across all types of fragments.
		These results are consistent with established design rules for acceptor materials. Indeed, the Scharber model estimates PCEs qualitatively from the electronic properties of the acceptor material.\cite{Scharber:2006, Ameri:2009} The design of non-fullerene acceptor candidates beyond the provided library could therefore be inspired mostly by the electronic properties of the fragments rather than their geometric properties, although more informative descriptors could potentially be constructed with more computational effort.\cite{Sahu:2018_opv}

	\subsection{Discovery of hybrid organic-inorganic perovskites for light harvesting}
	\label{sec:perovskites}

\begin{figure*}[htb]
			\centering
			\includegraphics[width = 1.0\textwidth]{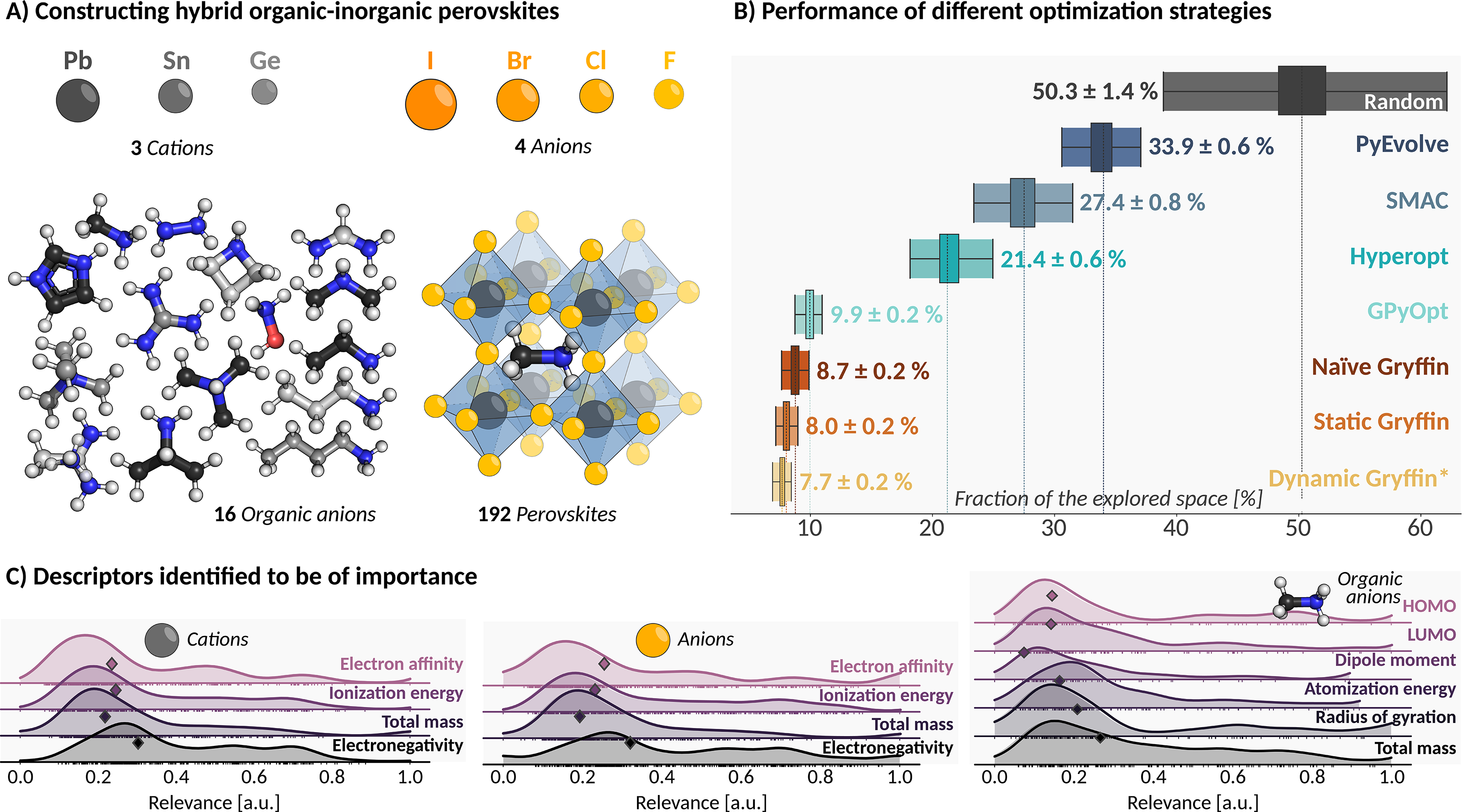}
			\caption{Results of the benchmarks on hybrid organic inorganic perovskites. (A) Perovskites are assembled by choosing one of three inorganic cations, one of four inorganic anions and one of 16 organic anions, resulting in $192$ unique designs. (B) Fractions of the candidate library to be explored by each of the studied optimization strategies to identify the perovskite design with the lowest bandgap. (C) Most informative descriptors to guide the optimization for each of the three constituents identified by dynamic \gryffin. The relevance metric is computed as the normalized weights that \emph{dynamic} \gryffin learnt to associate to each descriptor. The distributions shown are over the $200$ repeated optimizations performed. Diamonds indicate the average relevance of each descriptor.}
			\label{fig:app_org_inorg_results}
		\end{figure*}

		Perovskite solar cells constitute another class of light-harvesting materials. They are typically composed of inorganic lead halide matrices and contain inorganic or organic anions (see Fig.~\ref{fig:app_org_inorg_results}a).\cite{Jeon:2015_perovskite, Yang:2015_perovskite, Nie:2015_perovskite} Recently, perovskite solar cells have experienced increased attention as breakthroughs in materials and device architectures boosted their efficiencies and stabilities.\cite{Tan:2017_perovskite} The discovery of a viable perovskite design involves numerous choices regarding material compositions and process parameters, which poses a challenge to the rapid advancement of this light-harvesting technology. This second demonstration of the applicability of \gryffin focuses on the discovery of hybrid organic-inorganic perovskites (HOIPs) based on a recently reported dataset.\cite{Kim:2017} The HOIP candidates of this dataset are designed from a set of four different halide anions, three different group-IV cations and 16 different organic anions, resulting in 192 different HOIP compositions. Among other properties, this dataset reports the bandgaps of the HOIP candidates obtained from DFT calculations with GGA and the HSE06 functional. In this application, we aim to minimize the HSE06 bandgaps, which are considered to be more accurate with respect to experiment than the GGA values.\cite{Kim:2017}
		
		The inorganic constituents are characterized by their electron affinity, ionization energy, mass and electronegativity to guide the searches of the static and dynamic formulations of \gryffin. The organic compounds are described by their HOMO and LUMO energy levels, dipole moment, atomization energy, radius of gyration and molecular weight. All electronic descriptors were computed at the HSEH1PBE/Def2QZVPP level of theory on a SuperFineGrid with Gaussian,\cite{gaussian:2016} and the radii of gyration were calculated for the lowest-energy conformer. Note that, in contrast to the search for viable non-fullerene acceptors, this application presents an optimization task that not only features physically different descriptors between individual categorical variables, but also varying dimensionalities of the descriptors associated with individual categorical variables. However, the correlations between individual descriptors and the expected bandgaps of the assembled HOIP materials are significantly higher compared to the descriptors used for the non-fullerene acceptors (see supplementary information, Sec.~\ref{supp_sec:perovskites}, for additional details). 

		The fractions of the candidate space explored by each optimization strategy before locating the HOIP composition with the lowest bandgap are illustrated in Fig.~\ref{fig:app_org_inorg_results}b. More detailed results are reported in the supplementary information (see Sec.~\ref{supp_sec:perovskites}). Similarly to the synthetic benchmarks (see Sec.~\ref{sec:results}) and the optimization of non-fullerene acceptors (see Sec.~\ref{sec:non_fullerenes}), we find that all optimization strategies outperform a purely random exploration of the candidate space. Bayesian optimization strategies tend to locate the best performing HOIP candidate at a faster rate than \emph{PyEvolve} ($\sim\unit[34]{\%}$), with \emph{GPyOpt} evaluating only about \unit[10]{\%} of the candidate space, followed by \emph{Hyperopt} ($\sim\unit[21]{\%}$) and \emph{SMAC} ($\sim\unit[27]{\%}$). Na\"ive \gryffin succeeds after exploring less than \unit[9]{\%} of the search space. Note that this fraction of the search space corresponds to roughly 17 HOIP candidates, which approximately matches the number of available organic compounds.

		Despite na\"ive \gryffin already outperforming all other search strategies tested, its static and dynamic formulations still manage to improve upon \gryffin's performance and identify the best-performing HOIP after exploring less than \unit[8]{\%} of the design space, i.e. less than 16 HOIP candidates. This means that, on average, \gryffin could find the optimal design without the need to evaluate all organic anions. This observation further confirms that \gryffin's optimizations are indeed accelerated by the availability of suitable descriptors. However, in this example, we did not observe a significant performance difference between the static and the dynamic formulation of \gryffin. This behavior is expected given the high correlation (up to $0.9$) between provided descriptors and the bandgaps (see Sec.~\ref{sec:learning_descriptors}). 
		
		For this application, we find that electronegativity is most relevant for the inorganic constituents, while the radius of gyration and the molecular weight are most informative for the organic compound.
		Although the targeted property (bandgap of the HOIP) is an electronic property, dynamic \gryffin seems to benefit the most from the geometric (and not the electronic) descriptors of the organic compound. In contrast, the mass of the inorganic compounds seems to be the least relevant, while their electronegativity is most informative. These observations suggest that the organic molecule does not directly affect the electronic properties of the HOIP material, but rather induces a change in the arrangement of the inorganic compounds which in turn modulates the bandgap. Indeed, this hypothesis has emerged in various studies on perovskite materials.\cite{Kim:2016, Kim:2015, Umari:2014, Giorgi:2014, Endres:2016} These results further demonstrate how dynamic \gryffin is able to capture relevant and sometimes unexpected trends in the descriptors and inform future design choices.

	\subsection{Suzuki-Miyaura cross-coupling optimization}
	\label{sec:suzuki}

		As a final application, we demonstrate how \gryffin can aid in the optimization of Suzuki-Miyaura cross-coupling reactions with heterocyclic substrates.\cite{Miyaura:2004} These reactions are of particular interest to the pharmaceutical industry,\cite{Brown:2015} and have recently been studied in the context of self-optimizing reactors for flow chemistry.\cite{Baumgartner:2018, Reizman:2016, Reizman:2015} The optimization of chemical reactions typically targets a maximization of the yield. The yield of a reaction can be modified by varying its process conditions, which can largely be described by continuous variables. However, the reaction rate is also affected by the choice of catalyst, which is a categorical variable and cannot be described with a continuous value.

\begin{figure*}[htb]
			\includegraphics[width = 1.0\textwidth]{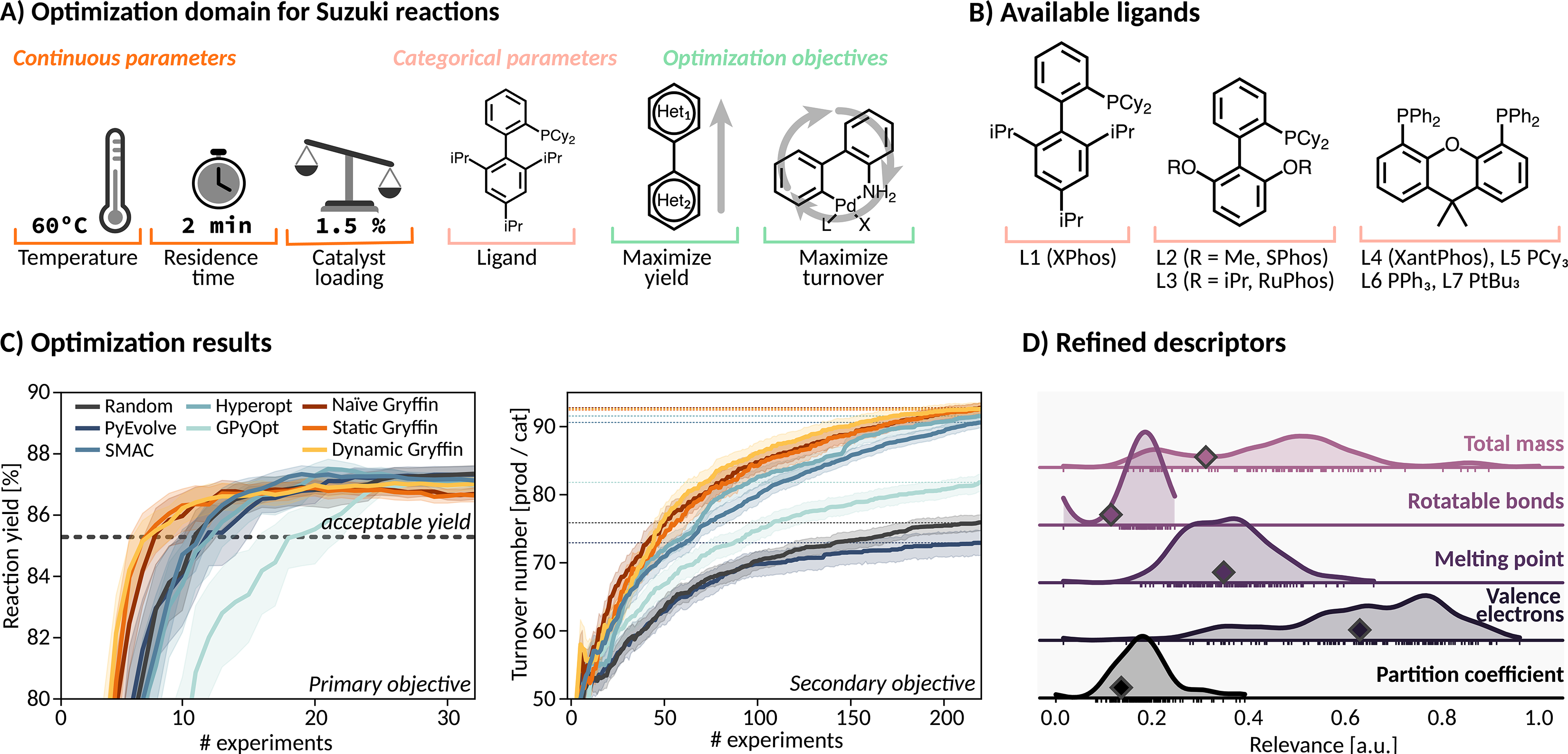}
			\caption{Results of the benchmarks on the optimization of Suzuki-Miyaura reactions. (A) Parameters and objectives for the optimization task: We target the identification of optimal values for three process parameters (temperature, residence time, catalyst loading) and one categorical variable (ligand) to maximize the yield of the reaction and the turnover number (TON) of the catalyst. (B) Illustration of the seven available ligands. (C) Optimization traces showing the performance of individual optimization strategies. (D) Most informative descriptors to guide the optimization. The relevance metric is computed as the normalized weights that \emph{dynamic} \gryffin learnt to associate to each descriptor. The distributions shown are over the $200$ repeated optimizations performed. Here, the relevance is indicative of the influence of each descriptor on both objectives. Diamonds indicate the average relevance of each descriptor.}
			\label{fig:suzuki_results}
		\end{figure*}

		Here, we consider a flow-based Suzuki-Miyaura cross-coupling reaction, in which we tune three continuous reaction conditions (temperature, residence time, catalyst loading) and one categorical variable (ligand for Palladium catalyst) as illustrated in Fig.~\ref{fig:suzuki_results}a. In this optimization task, we aim at maximizing both the reaction yield and the turnover number (TON) of the catalyst. We employ the \emph{Chimera} scalarizing strategy\cite{Hase:2018_chimera} to enable this multi-objective optimization, where we accept a \unit[10]{\%} degradation on the maximum achievable reaction yield to increase the TON as the secondary objective. This acceptable degradation corresponds to a desired reaction yield of above \unit[85.4]{\%}. \gryffin extends the \phoenics algorithm to simultaneously optimize categorical and continuous parameters. We consider a set of seven ligands (see Fig.~\ref{fig:suzuki_results}b), which are characterized by their molecular weight, the number of rotatable bonds, their melting points, the number of valence electrons, and their partition coefficients (\emph{logP}). Details on the physicochemical descriptors and the ranges for the continuous parameters are provided in the supplementary information (see Sec.~\ref{supp_sec:suzuki}). 

		As a complete performance analysis of the different optimization strategies is experimentally not tractable, we emulate noisy experimental responses with a probabilistic model trained on experimental data.\cite{Hase:2018_chimera, Hase:2020_olympus} Specifically, we train a Bayesian neural network to reproduce the reaction yield and TON of a previously reported flow-based reactor.\cite{Reizman:2016} Details on the data acquisition, model training and prediction accuracies are provided in the supplementary information (see Sec.~\ref{supp_sec:suzuki}). The excellent performance ($r^2 > 0.96$) of the model on unseen data for both the reaction yield and the TON indicates it is an accurate approximation of the experimental surface. With this experiment emulator, we execute $200$ independent optimization runs, each with $240$ evaluations, for each of the eight experiment planning strategies tested.

		Fig.~\ref{fig:suzuki_results}c illustrates the individual performance of each optimization strategies. We find that \emph{GPyOpt} requires about $18$ evaluations to identify reaction conditions achieving the desired reaction yield above $\unit[85.4]{\%}$. The other benchmark strategies, including random exploration, satisfy this first objective already after evaluating approximately 10--12 different conditions. Only the three formulations of \gryffin can locate desired reaction conditions at an even faster rates, requiring 7--8 evaluations. For the optimization of TON, the secondary objective, we observe that \emph{PyEvolve} is the slowest optimization strategy. In fact, random search starts outperforming \emph{PyEvolve} after about $100$ evaluations. Despite its relatively poor performance on the reaction yield, \emph{GPyOpt} maximizes the TON faster than random search. Yet, \emph{SMAC} and \emph{Hyperopt} achieve significantly higher TONs for any given number of evaluations. They are slightly outperformed only by \gryffin. In this application, we do not observe a significant difference in the performance of the three formulations of \gryffin. This result can be attributed to the fact that we only have one categorical variable with only seven options to choose from. Nevertheless, dynamic \gryffin achieves slighly higher TONs than static or na\"ive \gryffin. 

		The contributions of individual descriptors are illustrated in Fig.~\ref{fig:suzuki_results}d, where we find that the number of valence electrons shows the highest relevance among all descriptors to guide dynamic \gryffin, while the number of rotatable bonds is the least relevant. Indeed, the number of rotatable bonds correlates the least with the maximum and average reaction yields and TONs for any values of the other parameters (see supplementary information, Sec.~\ref{supp_sec:suzuki}), confirming that dynamic \gryffin correctly identifies non-informative descriptors within the given library of ligand candidates. The number of valence electrons also correlates strongly with the maximum achievable reaction yield for each of the ligands, confirming that this descriptor is highly informative to identify ligands that satisfy the reaction yield threshold. Melting point and molecular weight are identified as relevant likely due to their strong correlation with the number of valence electrons. Based on the indications of dynamic \gryffin, the design of more potent ligand candidates could be inspired by the number of valence electrons. In contrast to the previous applications, however, we consider a relatively small library of only seven ligands, such that the descriptor indications might not necessarily generalize well to larger libraries.

		Overall, across all three applications, we find that na\"ive \gryffin generally constitutes a competitive strategy for the optimization of categorical variables in chemistry and materials science, which tends to outperform state-of-the-art optimization strategies. Static \gryffin can accelerate the search with provided descriptors and navigate the search space more efficiently by exploiting descriptor-based similarities between individual options, thus efficiently leveraging domain knowledge. Dynamic \gryffin can accelerate the search even further by transforming provided descriptors to improve their relevance and inspire scientific insights. Finally, \gryffin integrates well with optimization strategies for continuous variables and thus enables the simultaneous optimization of mixed continuous-categorical parameter spaces.

	\section{Discussion}

	In this work we introduced \gryffin, an experiment planning strategy for the selection of categorical variables, such as functional molecules, catalysts, and materials. Similar to a recently introduced algorithm for optimization over continuous domains\cite{Hase:2018_phoenics}, \gryffin relies on Bayesian kernel density estimation for the construction of a surrogate model. To extend this formulation of Bayesian optimization to categorical spaces, \gryffin takes advantaged of a smooth approximation of categorical variables.\cite{Jang:2016,Maddison:2016} Furthermore, by locally transforming the metric of the optimization domain, \gryffin is able to leverage domain knowledge in the form of descriptors. This unique feature allows it to exploit the similarity of different options to more efficiently navigate categorical spaces. In addition, because the relevance of each descriptor for the property being optimized can be investigated, \gryffin can also inform future design choices and spark scientific insight.

	We compared the performance of \gryffin to state-of-the-art categorical optimization algorithms on a set of synthetic benchmark functions. Our benchmarks indicate that the na\"ive formulation of \gryffin, which does not use any descriptor information, is competitive with state-of-the-art strategies on \emph{pseudo} convex surfaces and outperforms them on all other surfaces. Descriptor-guided searches with static \gryffin identify global optima at significantly faster rates across all surfaces. Dynamic \gryffin, which attempts to construct a more informative set of descriptors, can accelerate the search even further, especially when correlations between the descriptors and the properties to be optimized are moderate, and for noisy response surfaces. Importantly, across all our tests, the dynamic formulation of \gryffin was never found to be considerably inferior to its static formulation.

	The capabilities of \gryffin were further demonstrated on three real-world applications across materials science and chemistry: (i) the discovery of non-fullerene acceptors for organic solar cells, (ii) the discovery of hybrid organic-inorganic perovskites for light harvesting, and (iii) the mixed categorical-continuous selection of ligands and reaction conditions for Suzuki-Miyaura cross-coupling reactions. \gryffin outperformed the other experiment planning strategies in all three applications. Static and dynamic \gryffin accelerated the searches even with moderately informative physicochemical descriptors. We further found that dynamic \gryffin was able to identify trends among descriptors that elucidate some of the prevalent phenomena giving rise to the properties of interest. This observations indicates that dynamic \gryffin has the potential to foster scientific understanding and encourage physical and chemical intuition for the studied systems.
	
	A question that naturally arises when using \gryffin is how to select the descriptors associated with specific chemical species. In fact, cheminformatics tools like RDKit\cite{rdkit:2006} and Mordred\cite{Moriwaki:2018} can compute several hundred descriptors. In principle, descriptor selection should be guided by domain knowledge, and only descriptors that are expected to be informative should be included. This is because the higher the correlation between descriptors and objective, the more efficient the optimization will be. In addition, ideally, the chosen descriptor should not carry redundant information (i.e., be highly correlated between each other). However, in practice, we found that \gryffin is able to take advantage of descriptors with Pearson correlation magnitudes as low as 0.1 (Fig.~\ref{fig:descriptor_learning}). Furthermore, the inclusion of up to $64$ redundant descriptors was not found to have a significant impact on the performance of the algorithm (Fig.~\ref{supp_fig:num_desc_dependence}). A downside of using a large number (e.g., hundreds) of descriptors is the higher probability of observing spurious correlations between descriptors and the objective being optimized. Spurious correlations might result in some uninformative descriptors being considered relevant by dynamic \gryffin. However, the impact of possible spurious correlations is expected to be small when informative descriptors are present, and the uninformative descriptors will be identified as such when more data is collected. Hence, while it is prudent to carefully select the descriptors used, \gryffin is expected to be robust against uninformative or redundant descriptors. If in doubt whether a certain descriptor is informative to the specific optimization task, we suggest to include rather than remove the descriptor and let dynamic \gryffin learn its true relevance from the data collected during the optimization.
	
	Several extensions and improvements to \gryffin are possible and will be considered in future work. While dynamic \gryffin learns the importance of each descriptor from data, the static formulation assumes equal importance among all descriptors provided by the user. A possible extension could allow the user to provide weights associated with each descriptor to reflect user knowledge of their relative importance. Another possible improvement is the use of unsupervised learning techniques, such as variational autoencoders\cite{GomezBombarelli:2018,Siivola2020,Tripp2020}, to automatically obtain continuous molecular representations to be used instead of or in conjunction with user-defined descriptors.
	
	One of the most computationally demanding aspects of \gryffin is the optimization of the non-convex acquisition function. Within acquisition optimization, the evaluation of the objective function is responsible for most of the cost, given that numerous evaluations are required. The use of zeroth-order optimization approaches, like evolutionary algorithms, might reduce this cost by avoiding the function evaluations required by first- and second-order methods to estimate gradients. In addition, the use of compositional solvers\cite{Grosnit2020,Tutunov:2020} might improve the performance of the acquisition optimization routine when proposing experiments in batches. Overall, improvements in the algorithms used to optimize the acquisition function might result in better performance and a more computationally efficient code.

	The use of power transforms to handle heteroscedastic noise has recently been found to improve the performance of Bayesian optimization based on Gaussian processes.\cite{Cowen-Rivers2020} Among other recently proposed approaches to handle this scenario, common in many experimental and computational experiments, is the use of surrogate models that are robust against this type of uncertainty.\cite{Griffiths2019} Analogous approaches could be implemented and tested in \gryffin as well.

\section{Conclusion}
	In summary, \gryffin constitutes a readily available strategy for the efficient selection of categorical variables across optimization tasks in science and engineering, and alleviates some of the immediate challenges to the versatile deployment of autonomous experimentation platforms. The demonstrated improvement upon state-of-the-art approaches thanks to the use of physicochemical descriptors constitutes a step towards effective, data-driven experiment planning guided by domain knowledge. We believe that \gryffin will enable the acceleration of scientific discovery, such as the search for promising molecules and materials. We invite the community to test and deploy it in expensive optimization tasks where similarities between categorical options can be defined. \gryffin is available for download on GitHub.\cite{github_repo}

	\vspace{8mm}
	\section*{Acknowledgments}

		The authors thank Melodie Christensen, Dr.~Pascal Friederich, Dr.~Gabriel dos Passos Gomes and Dr.~Daniel P. Tabor for valuable and insightful discussions. F.H. acknowledges financial support from the Herchel Smith Graduate Fellowship and the Jacques-Emile Dubois Student Dissertation Fellowship. M.A. is supported by a Postdoctoral Fellowship of the Vector Institute. R.J.H. gratefully acknowledges the Natural Sciences and Engineering Research Council of Canada (NSERC) for provision of the Postgraduate Scholarships-Doctoral Program (PGSD3-534584-2019). L.M.R and A.A.G were supported by the Tata Sons Limited - Alliance Agreement (A32391). A.A.G. would like to thank Dr.~Anders Fr{\o}seth for his support. This work relates to Department of Navy award (N00014-19-1-2134) issued by the Office of Naval Research. The United States Government has a royalty-free license throughout the world in all copyrightable material contained herein. Any opinions, findings, and conclusions or recommendations expressed in this material are those of the authors and do not necessarily reflect the views of the Office of Naval Research. All computations reported in this paper were completed on the Arran cluster supported by the Health Sciences Platform (HSP) at Tianjin University, P.R. China and the Odyssey cluster supported by the FAS Division of Science, Research Computing Group at Harvard University.


	\phantomsection\addcontentsline{toc}{section}{\refname}\putbib[main]
\end{bibunit}

\clearpage
\newpage

\begin{bibunit}[unsrt]

	\onecolumngrid
	\setcounter{subsection}{0}
	\section*{Supplementary Information}

\setcounter{section}{0}
\renewcommand\thesection{S.\arabic{section}}
\def\thefigure{S.\arabic{figure}}
\def\thetable{S.\arabic{table}}

\section{Formulating \gryffin}

	\noindent In the following, we detail the derivation of the three variants of \gryffin for the descriptor-less and descriptor-guided optimization of categorical variables.

\subsection{Deriving kernel based Bayesian optimization for categorical variables}
\label{supp_sec:naive_gryffin_derivation}

	The na\"ive \gryffin approach for the optimization of categorical variables follows the recently introduced \phoenics strategy, which combines Bayesian kernel density estimation with Bayesian optimization for continuous parameter domains.\cite{hase:2018_phoenics} \phoenics is based on the idea that kernel densities of evaluated parameters can indicate promising regions in the parameter space where the global optimum could (not) be located based on the response values observed for the evaluated parameters. The na\"ive \gryffin approach extends this idea to categorical parameter spaces.

	\phoenics models kernel densities on continuous domains with normal priors, where the parameters of the normal priors are sampled from a Bayesian neural network conditioned on the observed parameter points. The use of normal priors provide two main advantages. First, the locations and the precisions of the priors can be controlled independently, which means that priors can be fine-tuned based on the collected observations. Second, the priors can be reparameterized into deterministic functions of their parameters and stationary noise nodes, which accelerates the computations for inference and prediction in automatic differentiation frameworks.\cite{Kingma:2013}

	The generalization of \phoenics to categorical parameters requires kernel density priors that satisfy these two criteria. One potential candidate is the recently introduced concrete distribution\cite{Maddison:2016} (simultaneously introduced as Gumbel-Softmax)\cite{Jang:2016}. This distribution is defined on the $n$-dimensional simplex $\Delta^{n-1}$, defined by $\vect{z} \in \Delta^{n-1} = \{ \vect{z} \in \mathbb{R}^n | z_i \in [0, 1] \text{ and } \sum_{i=1}^n z_i = 1\}$, with probability mass
	\begin{align}
		p_{\vect{\pi},\tau}(\vect{z}) = \Gamma(n) \tau^{n-1} \prod\limits_{k=1}^n \left( \frac{ \pi z_k^{-\tau -1}}{\sum_{i=1}^n \pi z_i^{-\tau}} \right),
	\end{align}
	
	where $\Gamma(n) = (n-1)!$ is the gamma function, $\tau$ is a temperature parameter that controls the precision of the distribution, and $\vect{\pi}$ indicate the class probabilities. Samples $\vect{z}$ of the concrete distribution can be generated by sampling from a standard Gumbel distribution, i.e.  we draw i.i.d. samples $g_k \sim \text{Gumbel}(0, 1)$, and setting
	
	\begin{align}
		z_k = \frac{\exp\left( \big( \log \pi + g_k) / \tau \right)}{\sum_{i=1}^n \exp\left( \big( \log \pi + g_k) / \tau \right)}.
	\end{align}
	
	Note, that the parameters $\vect{\pi}$ are deterministic. Based on the kernel densities $p_k(\vect{z})$ and the associated observed responses $f_k$, where $k$ indicates the $k$-th measurement, we can construct the acquisition function
	
	\begin{align}
		\alpha(\vect{z}) = \frac{\sum_{k=1}^n f_k p_k(\vect{z}) + \lambda p_\text{uniform}(\vect{z})}{\sum_{k=1}^n p_k(\vect{z}) + p_\text{uniform}(\vect{z})},
	\end{align}
	
	where $\lambda$ is a sampling parameter that controls the degree of exploitation or exploration expressed by the acquisition function.\cite{hase:2018_phoenics} New parameter points for future evaluations are suggested based on the location of the global optimum of this acquisition function, where $\lambda$ tunes its bias towards exploration ($\lambda \ll 0$) or exploitation ($\lambda \gg 0$). The global optimum of the acquisition function is located based on random sampling with a local refinement of promising candidates using L-BFGS.\cite{Fletcher:1987}

\subsection{Enabling descriptor-guided optimizations on categorical spaces}
\label{supp_sec:static_gryffin_derivation}

	The static \gryffin approach extends the categorical optimization strategy detailed in Sec.~\ref{supp_sec:naive_gryffin_derivation} to cases where individual options of a categorical variable $\vect{z}$ can be associated to a set of descriptors $\vect{x}$, for which we can define a metric to measure their pairwise distances. For simplicity, we consider descriptors that are $m$-dimensional real-valued vectors, $\vect{x} \in \mathbb{R}^m$, for which we compute distances as the Euclidean norm after normalization to the unit hypercube $[0,1]^m$. We further assume that the descriptors are uniquely associated with individual options, such that no two options are described by the same descriptor. 

	While options for future evaluations could be selected directly from the real valued space $\mathbb{R}^m$ in which the descriptors are embedded, the efficient selection of candidates solely based on their descriptors is non-trivial for various reasons. For example, the geometry of the descriptors embedded in the real space is unknown and can be highly complex for a generic optimization task, which poses a challenge to defining domain boundaries. Furthermore, only specific points in the descriptor space correspond to realizable options of the categorical variable, such that the selection of the next candidate to evaluate can be ambiguous. To avoid these challenges, static \gryffin instead suggests new options based on kernel densities constructed on the categorical space, where the geometry of the space is known. Static \gryffin accounts for descriptor information by locally transforming the metric on the categorical space based on varying distances between individual descriptors. To this end, static \gryffin constructs a metric tensor for the categorical space from the descriptors associated with individual options. 

	\begin{figure}[!ht]
		\includegraphics[width = 0.4\textwidth]{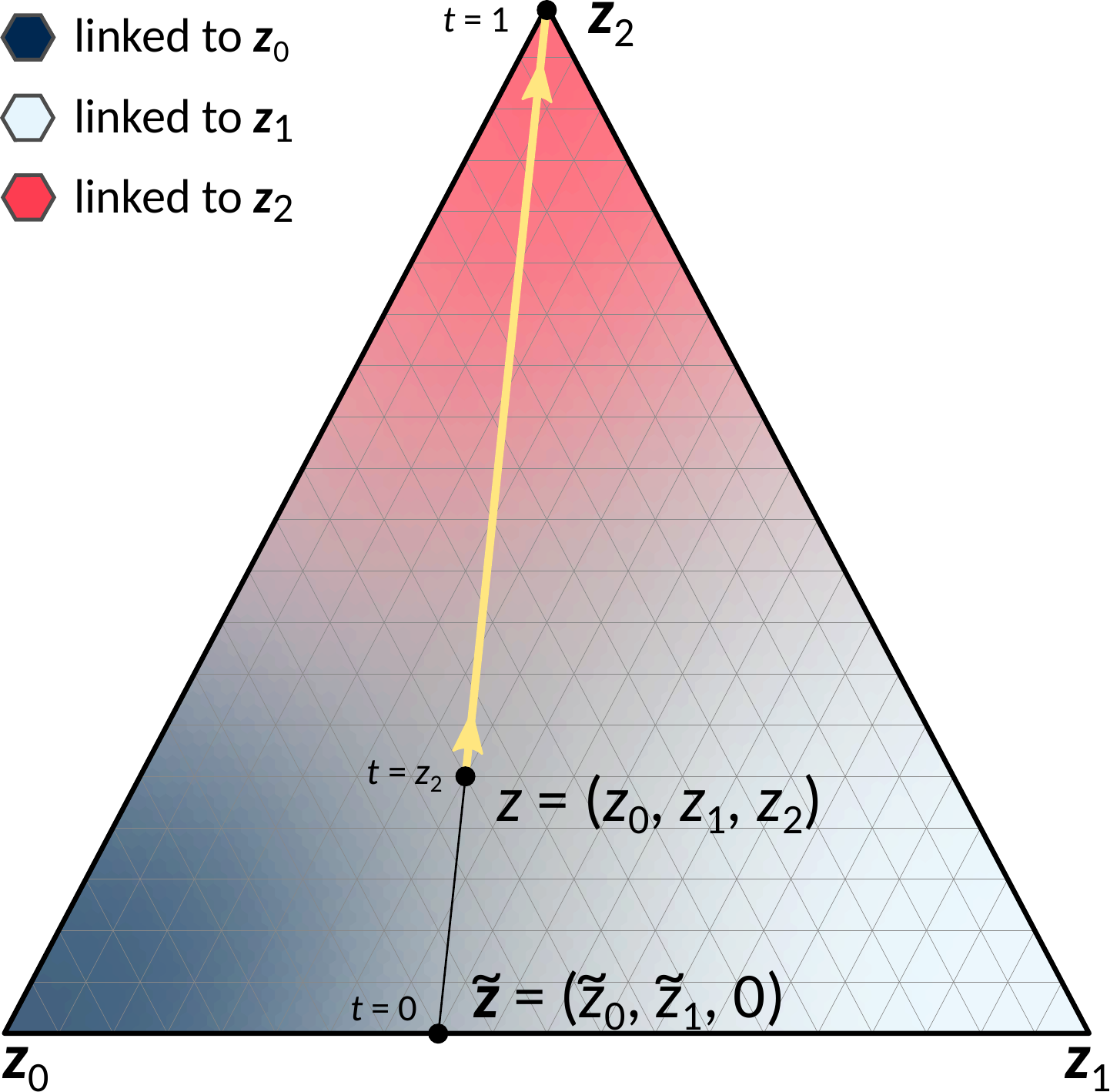}
		\caption{Illustration of the path in the categorical space along which changes of a given descriptor are computed for a change in a given categorical option. The path is parameterized by $t$ and goes from $\vect{z}$ to $\vect{z}_2$ in this example.}
		\label{supp_fig:similarity_construction}
	\end{figure}

	In the following, we derive an expression for the distance between any point $\vect{z}$ on a simplex with dimensionality $\text{\#opt}$ and a particular target corner $\vect{e}$, $\vect{z}_t = \delta_{it} \vect{e}_i$ for $i=0,\ldots,\text{\#opt}$, computed based on the metric spanned by the descriptors associated with individual options, following the example illustrated in Fig.~\ref{supp_fig:similarity_construction}. The infinitesimal line element on the descriptor space with $\text{\#desc}$-many descriptors can be calculated following the Euclidean norm
	\begin{align}
		ds^2 = \sum\limits_{m = 1}^{\# \text{desc}} dx_m dx_m,
	\end{align}
	where we sum over all descriptors. With the assumption that descriptors are a function of the points on the simplex, $\vect{x} = \vect{x}(\vect{z})$, we can compute the infinitesimal changes in the descriptors via infinitesimal changes in the categorical variable
	\begin{align}
		dx_m = \sum\limits_{i = 1}^{\# \text{opt}} \frac{\partial x_m}{\partial z_i} dz_i.
	\end{align}
	The infinitesimal line element $ds$ can then be expressed as
	\begin{align}\label{supp_eq:metric_tensor}
		ds^2 = \sum\limits_{m = 1}^{\# \text{desc}} \sum\limits_{i, j = 1}^{\# \text{opt}} \frac{\partial x_m}{\partial z_i} \frac{\partial x_m}{\partial z_j} dz_i dz_j
	\end{align}
	Further, if we assume that $\vect{x}(\vect{z})$ is a linear function of the points on the simplex, we can replace $dx \rightarrow \Delta x$, $dz \rightarrow \Delta z$ and $ds \rightarrow \Delta s$. The length of the path from a point $\vect{z}$ on the simplex to the corner $\vect{z}_i$ then simplifies to
	\begin{align}
		\Delta s^2 &= \sum\limits_{m = 1}^{\# \text{desc}} \sum\limits_{i, j = 1}^{\# \text{opt}} \frac{\Delta x_m}{\Delta z_i} \frac{\Delta x_m}{\Delta z_j} \Delta z_i \Delta z_j  = \sum\limits_{m = 1}^{\# \text{desc}} \sum\limits_{i, j = 1}^{\# \text{opt}} \Delta x_m^i\, \Delta x_m^j,
	\end{align}
	where we introduced $\Delta x_m^i$ to describe the change of the $m$-th descriptor with a change in the $i$-th option of the categorical variable. To compute this change in the descriptor, we consider a straight line in the categorical space as illustrated in Fig.~\ref{supp_fig:similarity_construction}.
	\begin{align}
		\vect{z}(t) = (1 - t) \tilde{\vect{z}} + t \vect{z}_i,\qquad\text{where} \quad t \in [0, 1]
	\end{align}
	and compute the value of the descriptor $\vect{x}_m$ along this path. For $t=1$, the value of $\vect{x}_m$ is identical to the value of the $m$-th descriptor of the $i$-th categorical option. For $t=0$, however, the value of $\vect{x}_m$ is given by the weighted average of the descriptors $x_m$ across all categories but the $i$-th category.
	\begin{align}
		x_m(t=0) = \sum\limits_{k\neq i}^{\# \text{opt}} \frac{z_k}{1 - z_i} x_m^{\text{opt}_k},\qquad x_m(t = 1) = x_m^{\text{opt}_i}
	\end{align}
	Hence, for a given point $\vect{z}$ along this path,
	\begin{align}
		x_m(\vect{z}) = z_i x_m^{\text{opt}_i} + (1 - z_i) \sum\limits_{k\neq i}^{\# \text{opt}} \frac{z_k}{1 - z_i} x_m^{\text{opt}_k}
	\end{align}
	where $x_m^{\text{opt}_i}$ denotes the value of the $m$-th descriptor of the $i$-th categorical option. Following the path from $\vect{z}$ to $\vect{z}_i$ (yellow line in Fig.~\ref{supp_fig:similarity_construction}), we find that $x_m$ changes as
	\begin{align}
		\Delta x_m^i = x_m^{\text{opt}_i} - \sum\limits_{k = 1}^{\# \text{opt}} z_k x_m^{\text{opt}_k}
	\end{align}
	We then compute the length of the path outlined in Fig.~\ref{supp_fig:similarity_construction} to arrive at
	\begin{align}\label{supp_eq:recomputed_distances}
		\Delta s^2 &= \sum\limits_{m = 1}^{\# \text{desc}} \sum\limits_{i, j = 1}^{\# \text{opt}} \left( x_m^{\text{opt}_i} - \sum\limits_{k = 1}^{\# \text{opt}} z_k x_m^{\text{opt}_k} \right) \left( x_m^{\text{opt}_i} - \sum\limits_{k = 1}^{\# \text{opt}} z_k x_m^{\text{opt}_k} \right) \notag \\
			&= (\# \text{opt})^2 \sum\limits_{m = 1}^{\# \text{desc}} \left( x_m^{\text{opt}_i} - \sum\limits_{k = 1}^{\# \text{opt}} z_k x_m^{\text{opt}_k} \right) \left( x_m^{\text{opt}_i} - \sum\limits_{k = 1}^{\# \text{opt}} z_k x_m^{\text{opt}_k} \right)
	\end{align}
	Eq.~\ref{supp_eq:recomputed_distances} presents the final equation to recompute distances. Based on these distances, similarity between sampled points on the simplex and its corners can be established. 

	\begin{figure}[!ht]
		\includegraphics[width = 0.8\textwidth]{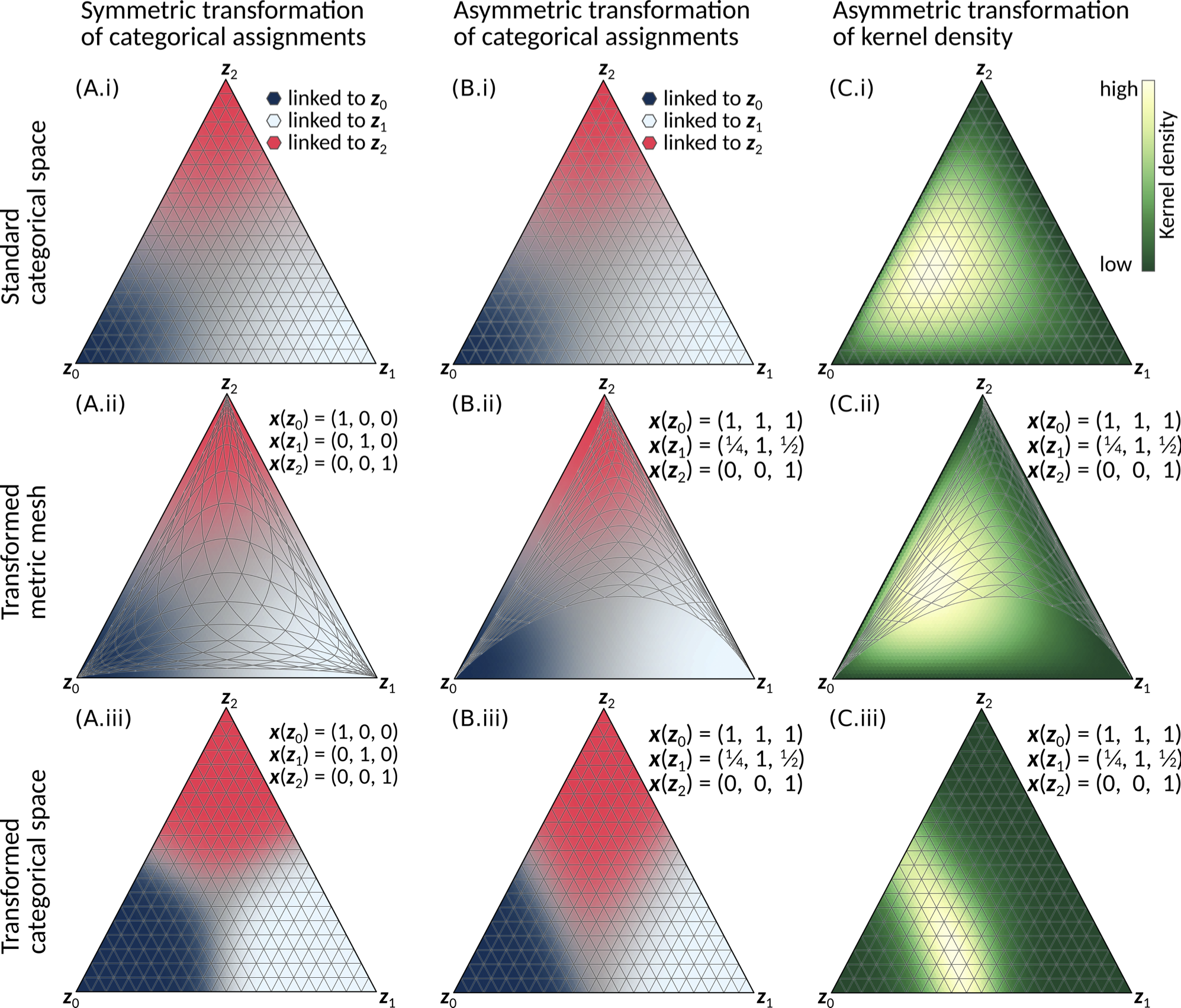}
		\caption{Illustration of a three dimensional categorical space, where a metric is defined based on the Euclidean norm between descriptors associated to the options of the categorical space (in comparison to a Euclidean metric on the categorical space). (A) Illustration of a descriptor-guided metric with equidistant descriptors. Color contours indicate the closest options. (B) Illustration of a descriptor-guided metric with arbitrary descriptors. Color contours indicate the closest option. (C) Illustration of the transformation of a kernel density on the categorical space based on the metric imposed by the descriptors used in panels (B). Color contours depict the kernel density. Panels (i) highlight contours and meshes on the simplex with a Euclidean norm. Panels (ii) illustrate meshes for the descriptor-guided metric on color contours for the Euclidean norm. Panels (iii) show color contours for the descriptor-guided metric with meshes represented in the Euclidean norm. }
		\label{supp_fig:space_reshaping_examples}
	\end{figure}

	The reshaping process is further illustrated on a three dimensional categorical space in Fig.~\ref{supp_fig:space_reshaping_examples} for two different sets of descriptors. Fig.~\ref{supp_fig:space_reshaping_examples}a is generated for a set of three dimensional descriptors with equal pairwise distances between them, with $x_0 = (1, 0, 0)$, $x_1 = (0, 1, 0)$ and $x_2 = (0, 0, 1)$. Fig.~\ref{supp_fig:space_reshaping_examples}b,c are generated for descriptors with different pairwise distances arising from $x_0 = (1, 1, 1)$, $x_1 = (1/4, 1, 1/2)$, and $x_2 = (0, 0, 1)$. Panels (i) are created with the Euclidean metric on the simplex, such that the color contours in panels A.i and B.i illustrate the closest corner point to any point in the simplex, and C.i shows an arbitrary kernel density sampled from a concrete distribution. The meshes illustrate line elements of equal length along individual coordinates. Panels (ii) show the meshes computed for the descriptor-guided metric, while contour plots follow the Euclidean metric. Panels (iii) show the contours for the descriptor-guided metric and the meshes for the Euclidean metric.

\subsection{Data-driven construction of more informative descriptors}
\label{supp_sec:dynamic_gryffin_derivation}

	Static \gryffin leverages real-valued descriptors to navigate categorical spaces in search for the best performing options. Such descriptor-guided searches, however, can only achieve faster optimization rates if the provided descriptors are representative of the expected performance of individual options. In fact, for an optimal performance of static \gryffin, the provided descriptors should perfectly correlate with the collected measurements, which will rarely be the case in real-world applications. However, descriptors that only poorly resemble the associated measurements cannot efficiently guide the search.

	The dynamic \gryffin approach aims to alleviate this limitation by constructing more informative descriptors, denoted with $\vect{x}^\prime$, on-the-fly from the responses $\{f_k\}$ collected during the optimization and the provided descriptors $\vect{x}$. The transformed descriptors $\vect{x}^\prime$ are considered to be more informative if they achieve higher correlations with the collected measurements $\{f_k\}$, which are quantified \emph{via} the Pearson correlation coefficient. Note, that we focus on large Pearson correlations without loss of generality, as highly negative coefficients close to $-1$ can be converted to large coefficients by a sign flip in the descriptors. The construction of a transformation $T$, which generates more informative descriptors $\vect{x}^\prime$ from the provided descriptors $\vect{x}$, i.e. $T: \vect{x} \mapsto \vect{x}^\prime$, can potentially elucidate the relevance of the provided descriptors, measured by their significance and influence to the construction of $\vect{x}^\prime$. As such, analyzing the transformation $T$ and identifying the descriptors $\vect{x}$ to which the measurements are most sensitive has the potential to inspire scientific insights.\cite{Hase:2019_interpretability} 

	Dynamic \gryffin implements the descriptor transformation $T$ targeting three goals.
	\begin{enumerate}[(i)]
		\item Maximizing the correlation between collected measurements $\{f_k\}$ and at least one of the transformed descriptors $\vect{x}_i^\prime$. Transformed descriptors that highly correlate with the collected measurements can efficiently guide an optimization strategy such as \gryffin, as demonstrated in Sec.~\ref{sec:results} of the main text.
		\item Reducing the cardinality of the set of transformed descriptors, such that only informative descriptors are retained. Larger sets of descriptors are more likely to span highly non-linear manifolds with stronger local curvatures, which can pose challenges to efficient optimizations.
		\item Reducing the pairwise correlation between any two transformed descriptors. Two highly correlated descriptors contain redundant information which is irrelevant for the optimization. The set of transformed descriptors is most informative if there is little correlation between any two transformed descriptors.
	\end{enumerate}

	\subsubsection{Defining the class of suitable transformations}

\begin{figure}[!ht]
			\includegraphics[width = 0.8\textwidth]{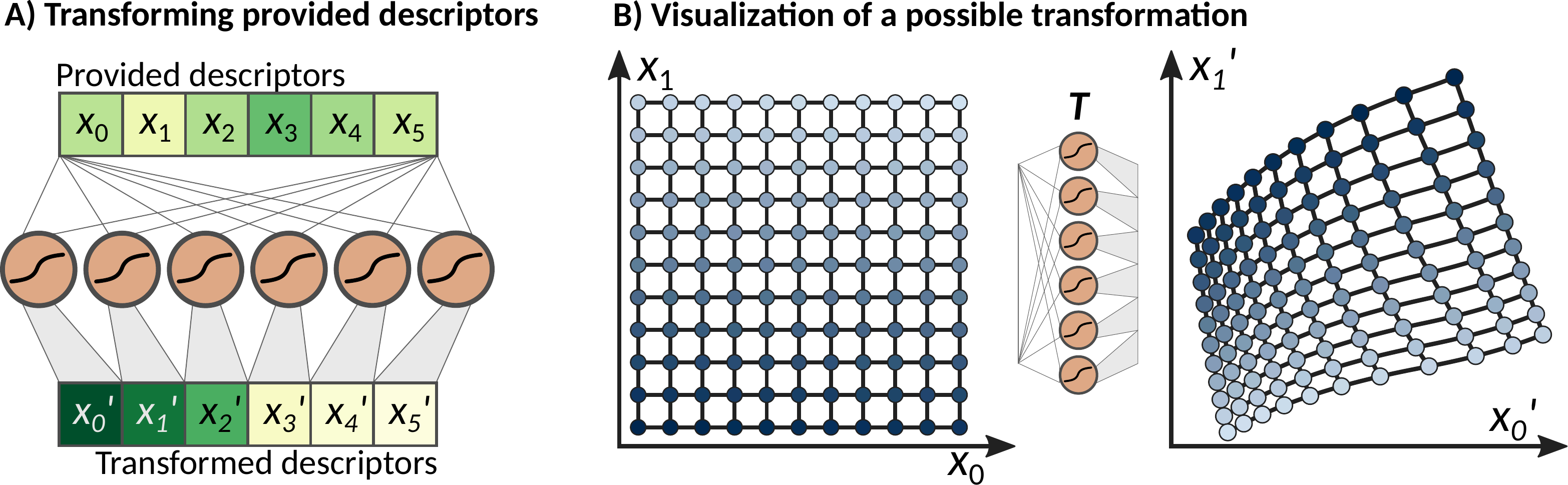}
			\caption{Illustration of the transformation $T$ leveraged by dynamic \gryffin to learn a more informative set of descriptors from collected measurements. (A) Architecture of the transformation, illustrated as a single-layer neural network. (B) Possible transformations spanned by the network architecture.}
			\label{supp_fig:learning_descriptors_explained}
		\end{figure}

		We model the transformation $T$ by a slightly non-linear operation, which could be interpreted as a simple, one-layer neural network. Specifically, we construct $T$ following Eq.~\ref{supp_eq:descriptor_transformation}, where $\vect{W}$ and $\vect{b}$ are trainable parameters inferred from the collected measurements.
		\begin{align}
		\label{supp_eq:descriptor_transformation}
			\vect{x}^\prime = T(\vect{x}; \vect{W}, \vect{b}) = \text{softsign}(\vect{W} \cdot \vect{x} + \vect{b}),\qquad\text{where}\quad\text{softsign}(x) = \frac{x}{1 + |x|}
		\end{align}
		This model architecture encodes slightly non-linear translations and rotations on the descriptor space. Although the simplicity of this model restricts the possible set of transformations, it substantially reduces the risk of overfitting in the low-data regimes commonly encountered in autonomous experimentation workflows. In addition, linear regression models are typically easier to interpret than more elaborate models with more complicated architectures, such that this choice of $T$ can potentially facilitate scientific insight. The architecture and the effect of realizable transformations on the descriptors are illustrated in Fig.~\ref{supp_fig:learning_descriptors_explained}.

	\subsubsection{Data-driven construction of suitable transformations}

		The parameters $\vect{W}$ and $\vect{b}$ of the transformation $T$ described by Eq.~\ref{supp_eq:descriptor_transformation} are inferred from collected measurements to satisfy the three aforementioned goals. To 3his end, we define a set of penalties, $\vect{\lambda}$, which are collectively minimized during inference via stochastic gradient descent. The first penalty, $\lambda_0$, targets the maximization of the correlation between at least one of $n$ transformed descriptors and the collected measurements
		\begin{align}
			\lambda_0 = 1 - \max\limits_{0 < i \leq n} |\rho(x_{i,k}^\prime, f_k)|,
		\end{align}
		where $\rho(x_{i,k}^\prime, f_k)$ denotes the Pearson correlation coefficient defined as
		\begin{align}
			\rho(x_{i,k}^\prime, f_k) = \frac{\mathbb{E}\left[ (x_{i,k}^\prime - \mu_{x^\prime}) (f_k - \mu_f) \right]}{\sigma_{x^\prime} \sigma_f},
		\end{align}
		and $\mu$ and $\sigma$ denote the mean and standard deviation of the descriptors and measurements over the set of all executed evaluations. Note, that this penalty favors correlation over anti-correlation without loss of generality. 
		
		Given at least one highly correlated descriptor, we would like the correlation between the measurements and all remaining descriptors to either be high (for the descriptors to be informative) or close to zero (for the descriptors to be insignificant). Additional, highly correlated descriptors $\vect{x}^\prime$ would still provide valuable information.
		Transformed descriptors with low correlations, however, can be neglected altogether as they provide little guidance to the optimizer. To construct a penalty that reflects these two desired outcomes, we first estimate the expected values for insignificant correlations. Given two independent random sequences with $n$ elements, the expected Pearson correlation coefficient $\rho$ is zero with a standard error $\Delta \rho$ of approximately
		\begin{align}
		\label{supp_eq:correlation_significance}
			\Delta \rho = \frac{1}{\sqrt{n - 3}},
		\end{align}
		as derived by Fisher.\cite{Fisher:1915, Fisher:1921} We consider a Pearson correlation to be significant if its magnitude is greater than the expected standard error. This choice corresponds to selecting correlations if they are associated with a p-value smaller than 0.32. We define the adjusted correlation $\tilde{\rho}$ as
		\begin{align}
			\tilde{\rho} = \max \left(\frac{|\rho| - \Delta \rho}{1 - \Delta \rho}, 0\right).
		\end{align}
		The second penalty is constructed from this adjusted correlation to equally favor correlations close to 1 or below the significance threshold
		\begin{align}
			\lambda_1 = \frac{1}{n} \sum\limits_{0 < i \leq n} \sin^2\left( \pi \tilde{\rho}(x_{i,k}^\prime, f_k) \right),
		\end{align}
		
		where $\pi$ here refers to the mathematical constant. Our third penalty aims to decorrelate the transformed descriptors $\vect{x}$, to diversify the information carried by each of them. Indeed, transformed descriptors that all perfectly correlate with the collected measurements also necessarily correlate with one another, and are thus redundant. In the definition of this third penalty, we again consider correlations to be insignificant if they are below the threshold defined in Eq.~\ref{supp_eq:correlation_significance}
		\begin{align}
			\lambda_2 = \frac{1}{n (n-1)}\sum\limits^n_{\substack{0 < i, j \leq n \\ i \neq j}} \sin^2\left( \frac{\pi}{2} \enskip \tilde{\rho}\left( x_{i,k}^\prime,  x_{j,k}^\prime\right) \right).
		\end{align}
		Finally, the parameters $\vect{W}$ of the transformation are $L1$-regularized to favor sparse operations which are easier to interpret. The regularization factor is denoted with $\nu$ and set to $10^{-3}$ in all experiments
		\begin{align}
			\lambda_3 = \nu \sum\limits_i | w_i |.
		\end{align}
		The overall penalty function
		\begin{align}
			\lambda = \lambda_0 + \lambda_1 + \lambda_2 + \lambda_3,
		\end{align}
		is then minimized via stochastic gradient descent until no significant improvement is observed over a period of $20$ epochs, or a total training duration of $1,000$ epochs has been reached.

\section{Synthetic benchmark functions}
\label{supp_sec:benchmark_functions}
\begin{figure}[!ht]
		\centering
		\includegraphics[width = 1.0\textwidth]{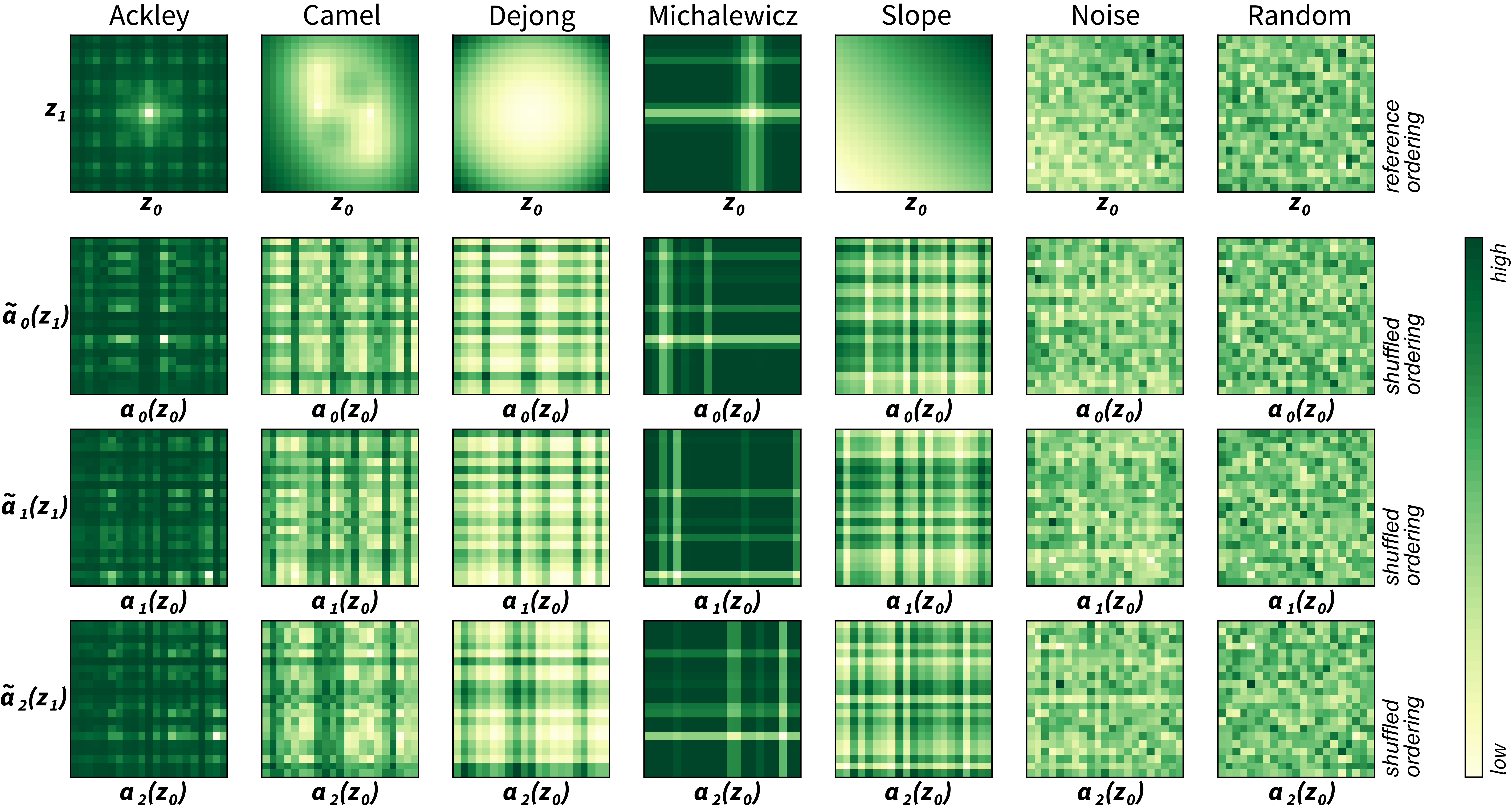}
		\caption{Synthetic benchmark functions generated for this study and employed for validating the introduced algorithm. The uppermost row depicts the benchmark functions in the \emph{reference} ordering of the options available to each variable. The lower rows illustrate the shape of the benchmark functions for \emph{shuffled} orderings generated from random permutations of the available options. Random permutations of the x-axis are indicated by $\alpha$, and $\widetilde{\alpha}$ indicates that different permutations have been applied to the y-axis.}
		\label{supp_fig:benchmark_functions}
	\end{figure}

	We demonstrate the performance of the introduced formulations of \gryffin with empirical benchmarks conducted on a set of synthetic surfaces. The benchmark surfaces are constructed with inspiration drawn from well-established analytic functions on continuous spaces that are typically used to benchmark local and global optimization strategies. Specifically, we extend the widely used \emph{Ackley}, \emph{Camel}, \emph{Dejong} and \emph{Michalewicz} functions to categorical spaces with arbitrarily many categorical variables and options per variable. In addition, we introduce three partially and fully randomized surfaces, referred to as \emph{Slope} (no noise), \emph{Noise} (moderate noise) and \emph{Random} (purely random), where responses are perturbed by noise sampled from stationary uniform distributions. All benchmark surfaces are illustrated in the top panels of Fig.~\ref{supp_fig:benchmark_functions} for two categorical variables with $21$ options per variable.

	It is important to note that there is no spatial relation between individual options since the order of any two options can be mutually switched without changing the surface. We illustrate this order ambiguity in the lower panels of Fig.~\ref{supp_fig:benchmark_functions}, where random permutations $\alpha$ have been performed on the \emph{reference} order illustrated in the top panels to create the \emph{shuffled} orders depicted in the lower panels. However, descriptors can be assigned to each of the categories such that a particular metric is imposed on the domain space. Unless noted otherwise, we use the \emph{reference} ordering for all options when supplying descriptors to construct a metric space for these categorical benchmark functions. 

	In the following, we describe and characterize the introduced benchmark surfaces in more detail. Python implementations of the benchmark functions are made available on GitHub.\cite{github_repo} 

	\paragraph{Ackley surface:} The \emph{Ackley} surface is inspired by the Ackley path function for continuous spaces. It features a narrow funnel around the global minimum, which is degenerate if the number of options along one (or more) dimensions is even and well-defined if the number of options for all dimensions is odd.

	\paragraph{Camel surface:} The \emph{Camel} surface is generalized from the Camel function on continuous domains and features a degenerate and \emph{pseudo}-disconnected global minimum.

	\paragraph{Dejong surface:} The \emph{Dejong} surface is inspired by the Dejong function and, as such, represents the generalization of a parabola to categorical spaces. We therefore refer to the \emph{Dejong} functions as \emph{pseudo}-convex. Similar to the \emph{Ackley} surface, the \emph{Dejong} surface features a well-defined global minimum if the number of options for all dimensions is odd, and a degenerate global minimum if at least one of the dimensions features an even number of options.

	\paragraph{Michalewicz surface:} The \emph{Michalewicz} surface is generalized to categorical spaces from the Michalewicz function. This surface features well-defined options for each dimension which yield significantly better performances than others. In addition, the number of \emph{pseudo}-local minima scales factorially with the number of dimensions.

	\paragraph{Slope surface:} The \emph{Slope} surface is constructed such that the response linearly increases with the index of the option along each dimension in the reference ordering. As such, the \emph{Slope} surface presents a generalization of a plane to categorical domains.

	\paragraph{Noise surface:} The \emph{Noise} surface is a variant of the \emph{Slope} surface, where perturbations sampled from a stationary uniform distribution are added to the responses of the surface, such that the overall correlation between the responses and the descriptors in the reference ordering target a Pearson correlation coefficient of $0.5$. The noise added to the surface is fixed with a random seed conditioned on the dimensionality and the number of options per dimension, such that the surfaces are generated reproducibly.

	\paragraph{Random surface:} The \emph{Random} surface is constructed from samples drawn from a stationary uniform distribution. To reproduce the surface, the random seed which generates the surface is conditioned on the dimensionality of the surface and the number of options per dimension.

\section{Empirical benchmarks of \gryffin on synthetic surfaces}
\label{supp_sec:synthetic_benchmarks}

	\noindent In the following sections we empirically illustrate the performance of the three introduced variants of \gryffin on the synthetic benchmark surfaces introduced in Sec.~\ref{supp_sec:benchmark_functions}.

	\subsection{Caching and boosting}
\label{supp_sec:boosting}

	The computationally most expensive step in the evaluation of the acquisition function is the computation of the kernel densities as an average over the number of samples drawn from the Bayesian neural network. However, the shape of the acquisition function is dominated by the (\emph{a priori} known) uniform distribution in regions of the parameter space where the kernel densities assume relatively low values. Based on this observation, we suggest that the construction of the acquisition function can be accelerated with an approximate scheme, which estimates kernel density values from fewer samples in low density regions.

	Following this strategy, we compute a first estimate to the value of the true kernel density $p_\text{true}(\vect{z})$ at a given parameter point $\vect{z}$ based on a randomly selected $\unit[10]{\%}$ of the samples drawn from the Bayesian neural network. This preliminary estimate, $p_{\text{approx},10}(\vect{z})$ is compared to the uniform distribution on the parameter domain, $p_\text{uniform}(\vect{z})$. If the estimated kernel density is greater than or equal to $\unit[1]{\%}$ of the uniform distribution, i.e.
	\begin{align}
		p_{\text{approx},10}(\vect{z}) \geq \frac{1}{100} \enskip p_\text{uniform}(\vect{z}),
	\end{align}
	the kernel density is considered to be sufficiently large to require a more accurate estimate, $p_{\text{approx},100}$, using $\unit[100]{\%}$ of the samples. Otherwise, the evaluation is stopped and the more uncertain estimate is used to approximate the true kernel density, thus saving $\unit[90]{\%}$ of the sample evaluations at this parameter point. Further accelerations of the implementation of the \gryffin framework are achieved by caching previously evaluated kernel densities to avoid redundant evaluations. This approach, however, balances reduced time requirements with slightly increased memory requirements and might thus not be applicable to all types of computational resources. We refer to this approximation as \emph{pseudo-boosting}.

\begin{figure}[!ht]
		\centering
		\includegraphics[width = 1.0\textwidth]{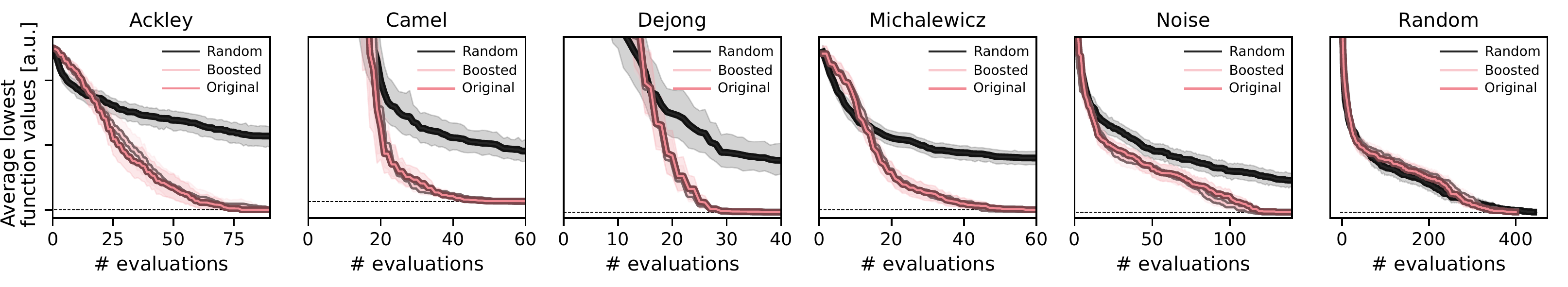}
		\caption{Performance of the pseudo-boosting and the full sampling formulations on six benchmark surfaces, averaged over $200$ independent repetitions. }
		\label{supp_fig:benchmark_boosting}
	\end{figure}

	We empirically estimate the runtime accelerations and the degree of potential performance degradations due to approximations to the acquisition functions on six of the benchmark surfaces introduced in Sec.~\ref{supp_sec:benchmark_functions}. The performances of the \emph{pseudo-boosting} strategy and the full sampling strategy are quantified based on the average best function values sampled for each benchmark surface after a certain number of iterations, while each optimization targets the location of the global minimum. Each benchmark surface is constructed with two categorical variables with $21$ options per categorical variable. Results for $200$ independent repetitions of the optimization runs for all surfaces are illustrated in Fig.~\ref{supp_fig:benchmark_boosting}. 

\begin{figure}
		\centering
		\includegraphics[width = 0.8\textwidth]{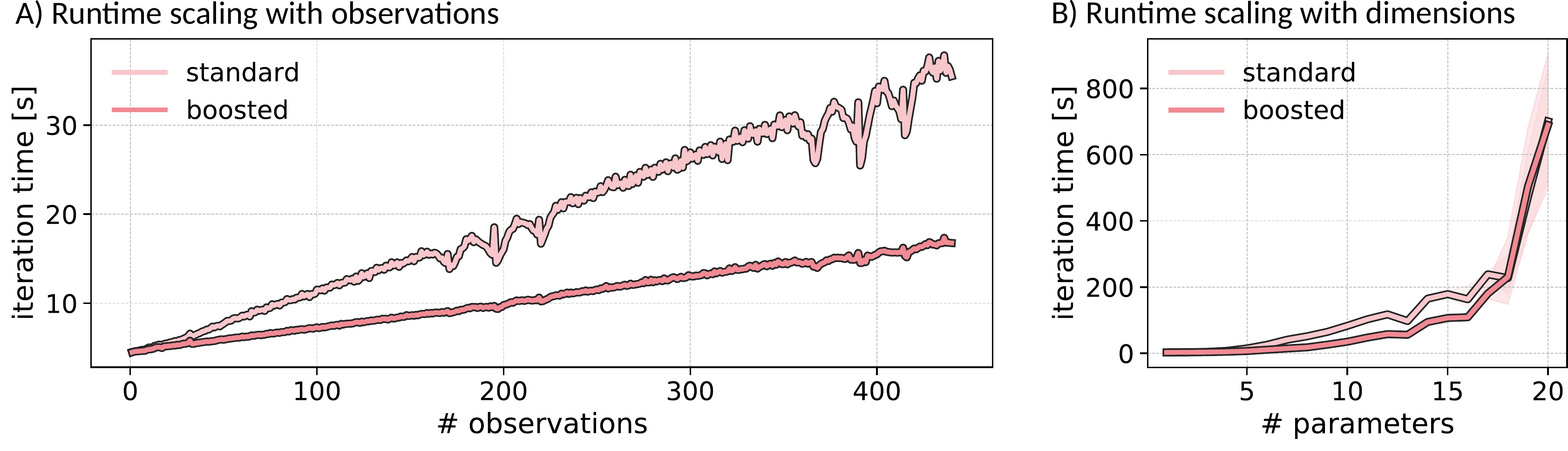}
		\caption{Illustrations of the computational scaling of \gryffin with and without \emph{pseudo-boosting}. (A) two dimensional response surface with varying number of observations. (B) ten observations with varying response surface dimensionality.}
		\label{supp_fig:benchmark_boosting_runtimes}
	\end{figure}

	The benchmarks indicate no significant performance difference between the \emph{pseudo-boosting} and the full sampling strategies of \gryffin, indicating that approximations to the acquisition function in low density regions does not severely affect the optimization runs. In addition to the performance, we also analyzed the computational time required for one iteration at a given number of options with fixed dimensionality (see Fig.~\ref{supp_fig:benchmark_boosting_runtimes}a), and varying number of parameters with a fixed number of options (see Fig.~\ref{supp_fig:benchmark_boosting_runtimes}a). All simulations were executed on an Intel(R) Core(TM) i5-7600K CPU at \unit[3.8]{GHz}.

	We find a linear dependence of the runtime with the number of observations in both cases. While the full sampling version of \gryffin experiences an increase in runtime of about \unit[0.071]{s} with each additional observation, the increase in runtime of the \emph{pseudo-boosted} version of \gryffin is about \unit[0.028]{s} per observation. Thus, the \emph{pseudo-boosted} implementation makes \gryffin's linear scaling $2.5$ times less steep. 
	However, we do not observe any significant accelerations for a varying number of dimensions. We conclude that \emph{pseudo-boosting} provides a runtime advantage when increasing the number of observations (as expected, by construction) without any noticeable degradations in the optimization performance. Based on these findings we recommend the \emph{pseudo-boosted} version of \gryffin and use the \emph{pseudo-boosted} version for all reported studies unless noted otherwise. 
	
	\subsection{Traces of the analytic benchmarks}
\label{supp_sec:analytic_benchmarks_traces}

\begin{figure}[!ht]
		\centering
		\includegraphics[width = 0.9\textwidth]{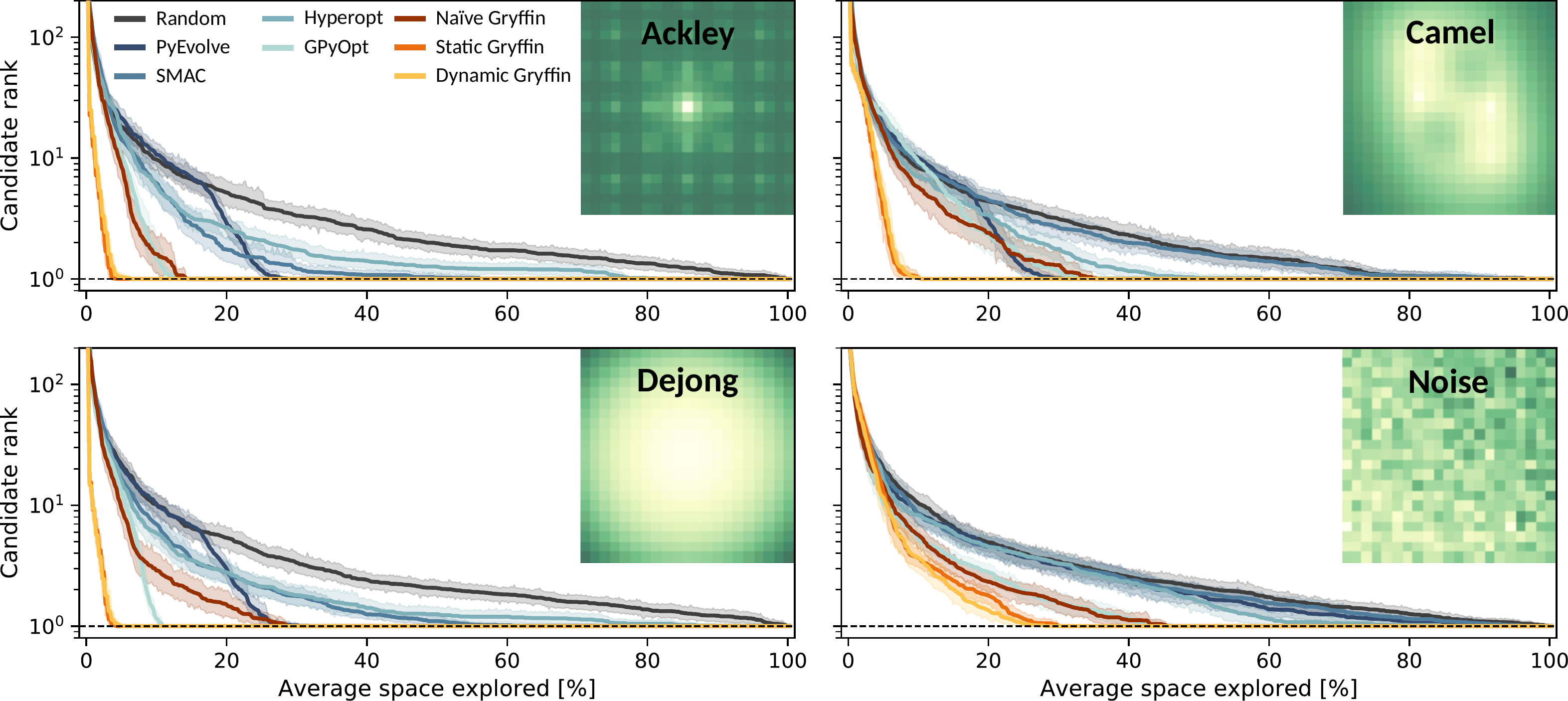}
		\caption{Average rank of the best performing candidate found by each of the studied optimization strategies in $200$ independent optimization runs. Benchmark surfaces are illustrated in the reference ordering for two categorical variables with $21$ options each. }
		\label{supp_fig:benchmark_traces}
	\end{figure}

	Fig.~\ref{supp_fig:benchmark_traces} illustrates the traces of $200$ independent optimization runs of each of the studied optimization strategies on four synthetic benchmark surfaces, supplementing the results reported in Sec.~\ref{sec:results} of the main text. Descriptors were provided in the reference ordering introduced in Sec.~\ref{supp_sec:benchmark_functions}.

\subsection{Influence of the number of dimensions}
\label{supp_sec:analytic_benchmarks_dimensions}

\begin{figure}[!ht]
		\centering
		\includegraphics[width = 1.0\textwidth]{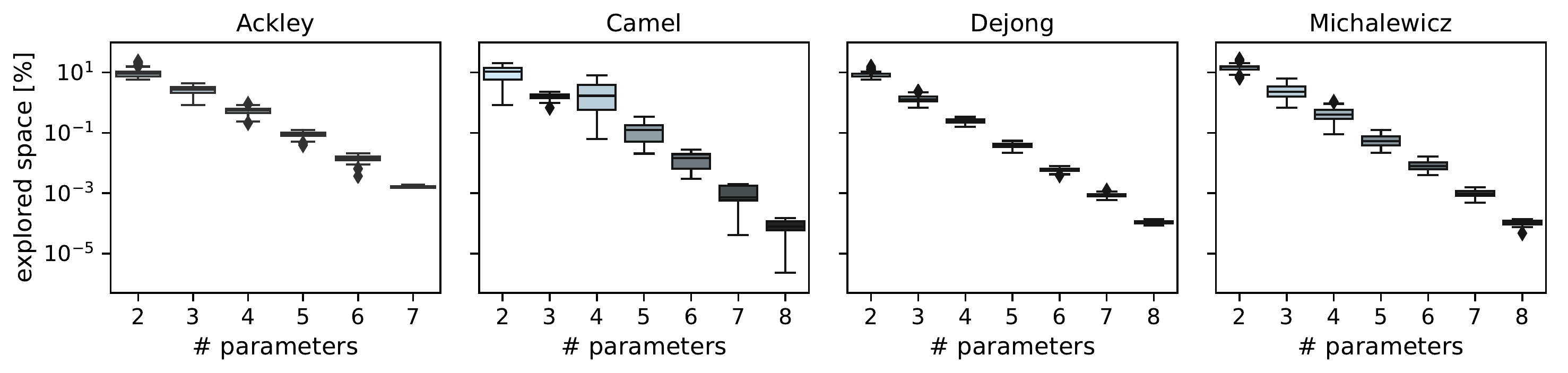}
		\caption{Fraction of explored space required by static \gryffin to locate the optimum for varying dimensionalities of different benchmark surfaces.}
		\label{supp_fig:num_dimensions}
	\end{figure}

	We empirically estimate the fraction of the candidate space that static \gryffin explores to locate the optimum with respect to the number of categorical variables in the optimization task. For this benchmark, we set up each categorical variable with a total of $11$ options and gradually increase the dimensionality of the search space. Fig.~\ref{supp_fig:num_dimensions} illustrates the performance of static \gryffin on different benchmark surfaces with varying dimensionality, for a total of $100$ independent executions. Descriptors were provided in the reference ordering introduced in Sec.~\ref{supp_sec:benchmark_functions}. 

	As expected, \gryffin required more candidate evaluations to locate the global optimum for search spaces with larger volumes. However,  the fraction of the space explored by static \gryffin decreased with the dimensionality across all benchmark surfaces. More specifically, we find that the dependence of the explored space to the number of parameters can be approximated with an exponential decay of the form $y = \alpha \exp(-\gamma x)$ where $y$ denotes the fraction of the explored space, $x$ denotes the number of parameters and $\alpha$ and $\gamma$ are two parameters which we infer from a least-square fit. Parameter values are reported in Tab.~\ref{supp_tab:dim_dependency}. This observation indicates that \gryffin does not suffer from the \emph{curse of dimensionality}, as the number of evaluated candidates does not increase as fast as the volume of the space.

	\begin{table}[!ht]
		\centering
		\begin{tabular}{lrrr}		         
			Surface     & $\alpha$ & $\gamma$ & $r^2$ score\\
			\hline
			Ackley      & 3.34     & 1.66     & 0.84   \\
			Camel       & 5.26     & 1.83     & 0.67   \\
			Dejong      & 3.86     & 1.86     & 0.95   \\
			Michalewicz & 8.17     & 1.95     & 0.90   \\
		\end{tabular}
		\caption{Fitting parameters for the dependency of the fraction of the explored space on the number of categorical parameters. The $r^2$ score indicates the coefficient of determination}
		\label{supp_tab:dim_dependency}
	\end{table}

\subsection{Influence of the number of options}
\label{supp_sec:analytic_benchmarks_options}

\begin{figure}[!ht]
		\centering
		\includegraphics[width = 1.0\textwidth]{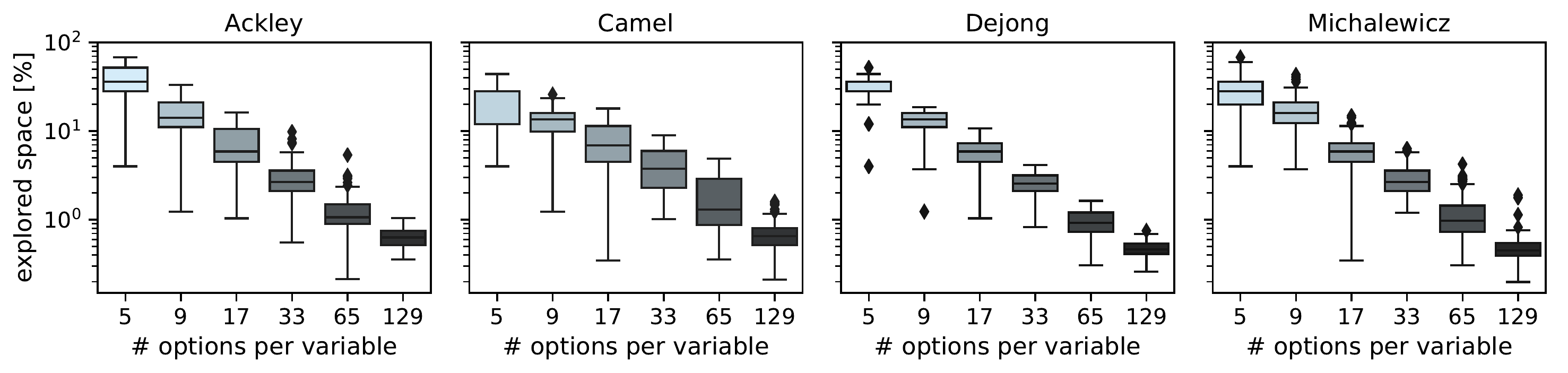}
		\caption{Fraction of explored space required by static \gryffin to locate the optimum depending on the number of options per parameter for different benchmark surfaces. Note, that the abscissa is not linear.}
		\label{supp_fig:num_options}
	\end{figure}

	We empirically estimate the fraction of the candidate space that static \gryffin explores to locate the optimum with respect to the number of options per categorical variable. We set up each benchmark surface with two categorical variables and gradually increase the number of options. Fig.~\ref{supp_fig:num_options} illustrates the performance of static \gryffin on different benchmark surfaces with varying options for a total of $100$ independent executions. Descriptors were provided in the reference ordering introduced in Sec.~\ref{supp_sec:benchmark_functions}. 

	We observe that the fraction of the space that static \gryffin explores to locate the optimum consistently decreases with an increasing number of options across all benchmark surfaces. More specifically, we find that the dependence of the explored space to the number of parameters can be approximated with an exponential decay of the form $y = \alpha x^{-\gamma}$ where $y$ denotes the fraction of the explored space, $x$ denotes the number of options and $\alpha$ and $\gamma$ are two parameters which we infer from a least-square fit. Parameter values are reported in Tab.~\ref{supp_tab:option_dependency}.

	\begin{table}[!ht]
		\centering
		\begin{tabular}{lrrr}
			Surface     & $\alpha$ & $\gamma$ & $r^2$ score\\
			\hline
			Ackley      & 2.34     & 1.25     & 0.73   \\
			Camel       & 1.20     & 1.05     & 0.68   \\
			Dejong      & 2.04     & 1.27     & 0.89   \\
			Michalewicz & 2.12     & 1.25     & 0.70   \\
		\end{tabular}
		\caption{Fitting parameters for the dependency of the fraction of the explored space on the number of categorical parameters. The $r^2$ score indicates the coefficient of determination}
		\label{supp_tab:option_dependency}
	\end{table}

\subsection{Influence of the number of descriptors}
\label{supp_sec:influence_of_descriptors}

	Static \gryffin facilitates the acceleration of categorical optimization by supplying an arbitrary number of real valued descriptors for every option of the categorical variable. This benchmark investigates the performance of \gryffin when changing the number of descriptors that construct the same metric space, while keeping the information content of the descriptors constant.  Simulations are run on the \emph{Michalewicz} surface (see Sec.~\ref{supp_sec:benchmark_functions}) with two categorical variables and $21$ options per variable. Descriptors are constructed for the reference ordering and repeated multiple times to span the same metric space. Optimization runs are repeated $100$ times with different random seeds to marginalize performance fluctuations.

	\begin{figure}[!ht]
		\centering
		\includegraphics[width=0.9\textwidth]{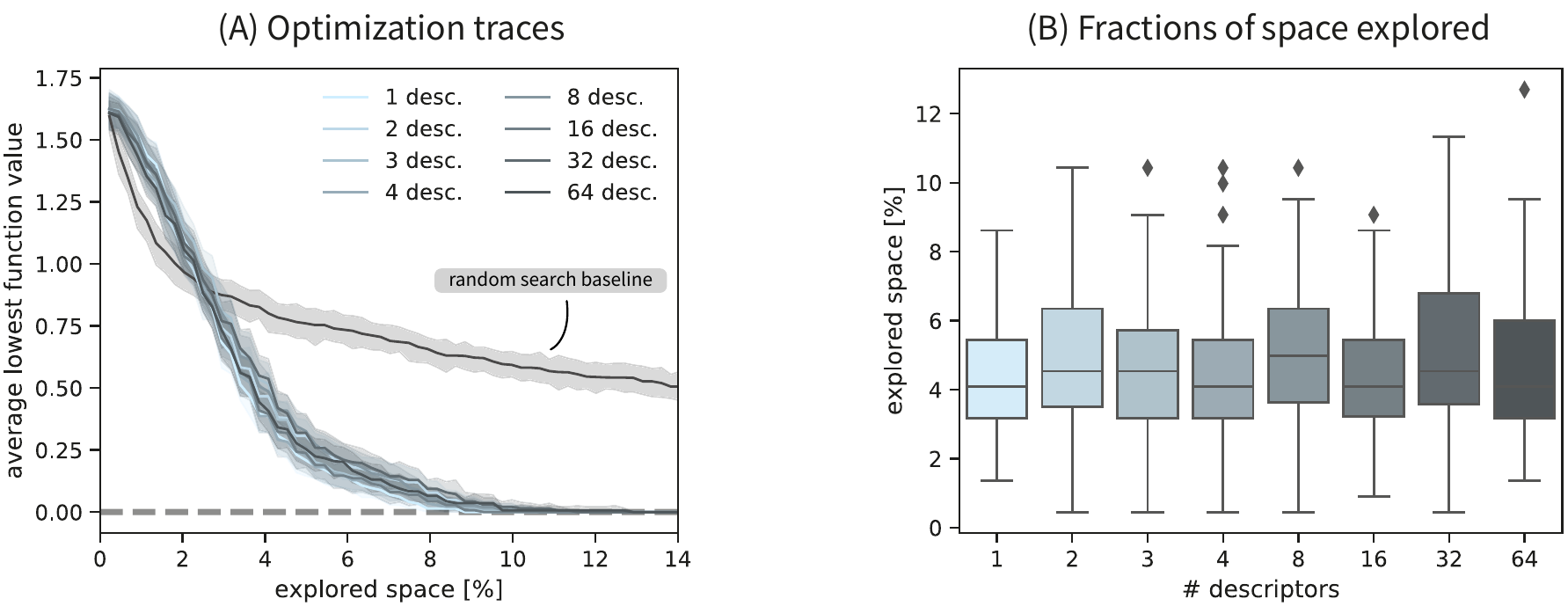}
		\caption{Performance of \gryffin with varying number of descriptors spanning the same metric space shown on the \emph{Michalewicz} surface in reference ordering. Optimization runs have been repeated $100$ times with different random seeds. (A) Achieved best function values in dependence of the explored fraction of the search space. The performance of a random search is shown for reference. (B) Fraction of the search space that had been evaluated when the optimal parameter combination was detected for different numbers of descriptors.}
		\label{supp_fig:num_desc_dependence}
	\end{figure}

	Overall, we find very similar performances across the different numbers of descriptors. Both the explored fraction of the space when detecting the optimum (Fig.~\ref{supp_fig:num_desc_dependence}b) and the traces of best values achieved during the optimization (Fig.~\ref{supp_fig:num_desc_dependence}a) do not display any significant differences. We conclude that the performance of \gryffin is not negatively affected by the presence of descriptors carrying redundant information.

\section{Real-world applications of \gryffin}

	\noindent Following the empirical benchmarks of \gryffin on synthetic functions (see Sec.~\ref{supp_sec:synthetic_benchmarks}) we demonstrate its applicability on three real-world examples: (i) the discovery of non-fullerene acceptors for organic solar cells, (ii) the discovery of hybrid organic-inorganic perovskites, and (iii) the optimization of reaction conditions for Suzuki-Miyaura cross-coupling reactions. Datasets for all three applications are made available on GitHub.\cite{github_repo}

	\subsection{Non-fullerene acceptors}
\label{supp_sec:non_fullerene_results}

	This discovery task involves the selection of molecular fragments that are assembled into non-fullerene acceptor candidates for solar cell applications, following a recent study by Lopez \emph{et al.}\cite{Lopez:2017} We consider a subset of the publicly available library with 4,216 acceptor candidates assembled from the following fragments:
	\begin{itemize}
		\item Terminal fragments: \texttt{frag\_31}, \texttt{frag\_32}, \texttt{frag\_33}, \texttt{frag\_34}, \texttt{frag\_36}, \texttt{frag\_38}, \texttt{frag\_47}, \texttt{frag\_49}, \texttt{frag\_51}, \texttt{frag\_52}, \texttt{frag\_63}, \texttt{frag\_64}, \texttt{frag\_68}, \texttt{frag\_72}, \texttt{frag\_114}, \texttt{frag\_115}, \texttt{frag\_119}
		\item Core fragments: \texttt{frag\_1}, \texttt{frag\_2}, \texttt{frag\_3}, \texttt{frag\_23}, \texttt{frag\_88}, \texttt{frag\_98}, \texttt{frag\_107}, \texttt{frag\_109}
		\item Spacer fragments: \texttt{frag\_4}, \texttt{frag\_5}, \texttt{frag\_6}, \texttt{frag\_7}, \texttt{frag\_14}, \texttt{frag\_17}, \texttt{frag\_19}, \texttt{frag\_22}, \texttt{frag\_24}, \texttt{frag\_25}, \texttt{frag\_46}, \texttt{frag\_55}, \texttt{frag\_57}, \texttt{frag\_58}, \texttt{frag\_60}, \texttt{frag\_61}, \texttt{frag\_81}, \texttt{frag\_82}, \texttt{frag\_85}, \texttt{frag\_90}, \texttt{frag\_100}, \texttt{frag\_101}, \texttt{frag\_105}, \texttt{frag\_108}, \texttt{frag\_110}, \texttt{frag\_112}, \texttt{frag\_120}, \texttt{frag\_121}, \texttt{frag\_127}, \texttt{frag\_128}, \texttt{frag\_129}
	\end{itemize}

	All fragments are characterized by a set of three electronic properties (HOMO and LUMO levels and the dipole moment), which were computed at the B3LYP/Def2SVP level of theory on a SuperFineGrid using Gaussian,\cite{gaussian:2016}, and two geometric properties (molecular weight and radius of gyration), which were computed for the ground state geometry. The power conversion efficiency (PCE) is the target of the optimization task, and the PCEs for all 4,216 candidates have already been computed and tabulated by Lopez \emph{et al.}\cite{Lopez:2017}. We can thus compute the correlation of every descriptor with the PCE. Results are reported in Tab.~\ref{supp_tab:nonfullerenes_correlations}, where we find that the electronic properties collectively correlate best with the optimization target (although correlations are relatively poor overall).

	\begin{table}[!ht]
		\centering
		\begin{tabular}{lrrr}
		 	& Core & Spacer & Terminus \\
			\hline
			HOMO               &  0.073          & -0.022          & \textbf{-0.080}  \\
			LUMO               & \textbf{-0.276} & -0.171          & -0.075           \\
			Dipole moment      & -0.220          &  \textbf{0.199} & -0.019           \\
			Molecular weight   &  0.199          & -0.003          & -0.056           \\
			Radius of gyration &  0.075          & -0.038          &  0.048           \\
		\end{tabular}
		\caption{Correlations of physical descriptors with observed performances of individual choices for termini, spacer and cores. Highest correlations are printed in bold.}
		\label{supp_tab:nonfullerenes_correlations}
	\end{table}

\begin{figure}[!ht]
		\includegraphics[width = 1.0\textwidth]{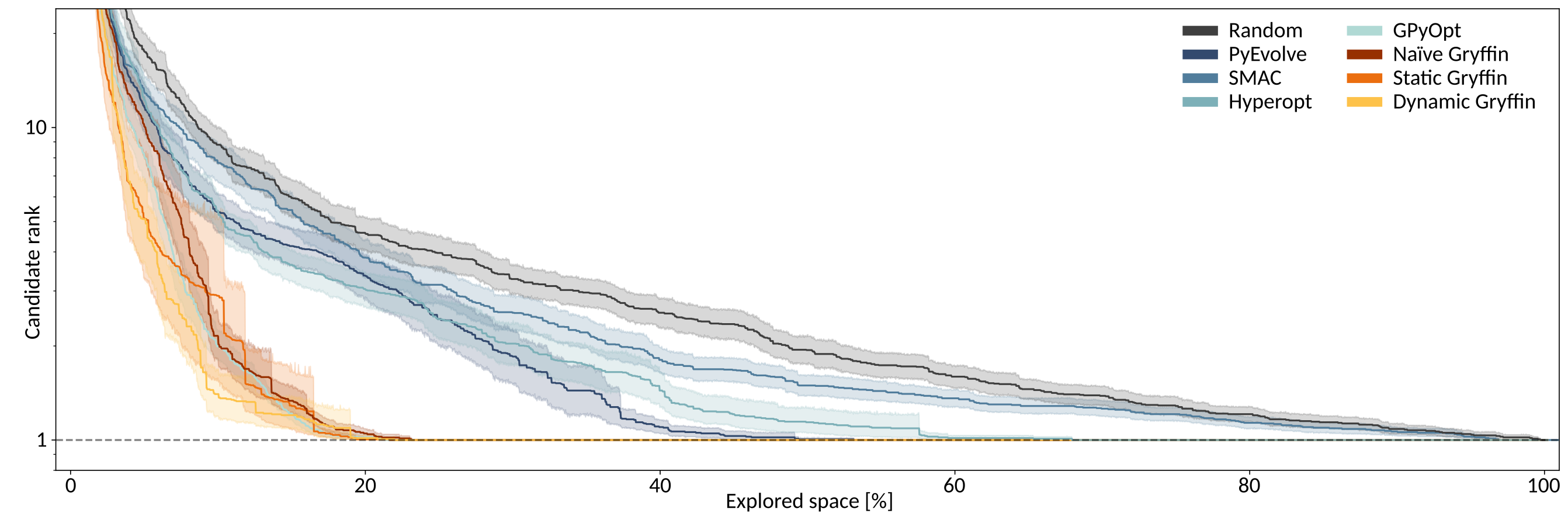}
		\caption{Average rank of the candidate with the highest power conversion efficiency identified by the different optimization strategies during $100$ independent optimization runs. }
		\label{supp_fig:non_fullerene_results}
	\end{figure}

	Fig.~\ref{supp_fig:non_fullerene_results} illustrates the performances of the studied optimization strategies on the non-fullerene application described in detail in Sec.~\ref{sec:non_fullerenes}. Each candidate molecule is assigned a rank based on its power conversion efficiency, starting from 1 for the candidate with the highest PCE, and up to 4,216 for the candidate with the lowest PCE. The graphs shown in Fig.~\ref{supp_fig:non_fullerene_results} illustrate the average rank of the best performing candidate for different stages of $200$ independent optimization runs.

	In agreement with the results reported in Sec.~\ref{sec:non_fullerenes}, we observe that the dynamic formulation of \gryffin locates promising candidates the fastest. While static \gryffin initially also shows a promising optimization rate, identifying candidates in the top 10 ranks with less than \unit[10]{\%} of the space explored, it noticeably slows down compared to the dynamic and even the na\"ive formulations, indicating that the provided descriptors might not directly favor the best performing candidate. This observation agrees with the fact that the descriptors overall correlate poorly with the optimization target. 
	
	\subsection{Hybrid organic inorganic perovskites}
\label{supp_sec:perovskites}

	We construct hybrid organic-inorganic perovskites (HOIP) from a set of different options for the inorganic cation, the inorganic anion, and the organic anion. HOIP designs are evaluated based on their bandgap, and ranked from lowest to highest. Inorganic constituents are characterized by their electron affinity, ionization energy, total mass and electronegativity. Organic anions are described by a set of electronic properties (HOMO, LUMO, dipole moment, atomization energy) and geometric properties (radius of gyration, molecular weight). Correlations between individual descriptors and the computed HOIP bandgaps are reported in Tab.~\ref{supp_tab:perovskite_correlations}

	\begin{table}[!ht]
		\centering
		\begin{tabular}{cc}
			\begin{minipage}{0.5\textwidth}
				\begin{tabular}{lrr}
			 		& Inorganic anion & Inorganic cation \\
					\hline
					Electron affinity &  0.452 & -0.116  \\
					Ionization energy &  \textbf{0.904} &  0.121  \\
					Total mass        & -0.804 &  0.069  \\
					Electronegativity &  0.902 &  \textbf{0.142}  \\
				\end{tabular}
			\end{minipage}
			&
			\begin{minipage}{0.5\textwidth}
				\begin{tabular}{lr}
					& Organic anion \\
					\hline
					HOMO               &  0.002 \\
					LUMO               &  0.140 \\
					Dipole moment      &  0.077 \\
					Atomization energy & \textbf{-0.159} \\
					Radius of gyration &  0.145 \\
					Molecular weight   &  0.138 \\
				\end{tabular}
			\end{minipage}
		\end{tabular}
		\caption{Correlations of physical descriptors with observed performances of individual choices of inorganic cation, inorganic anion, and organic anion. Largest correlation magnitudes are printed in bold.}
		\label{supp_tab:perovskite_correlations}
	\end{table}

\begin{figure}[!ht]
		\includegraphics[width = 1.0\textwidth]{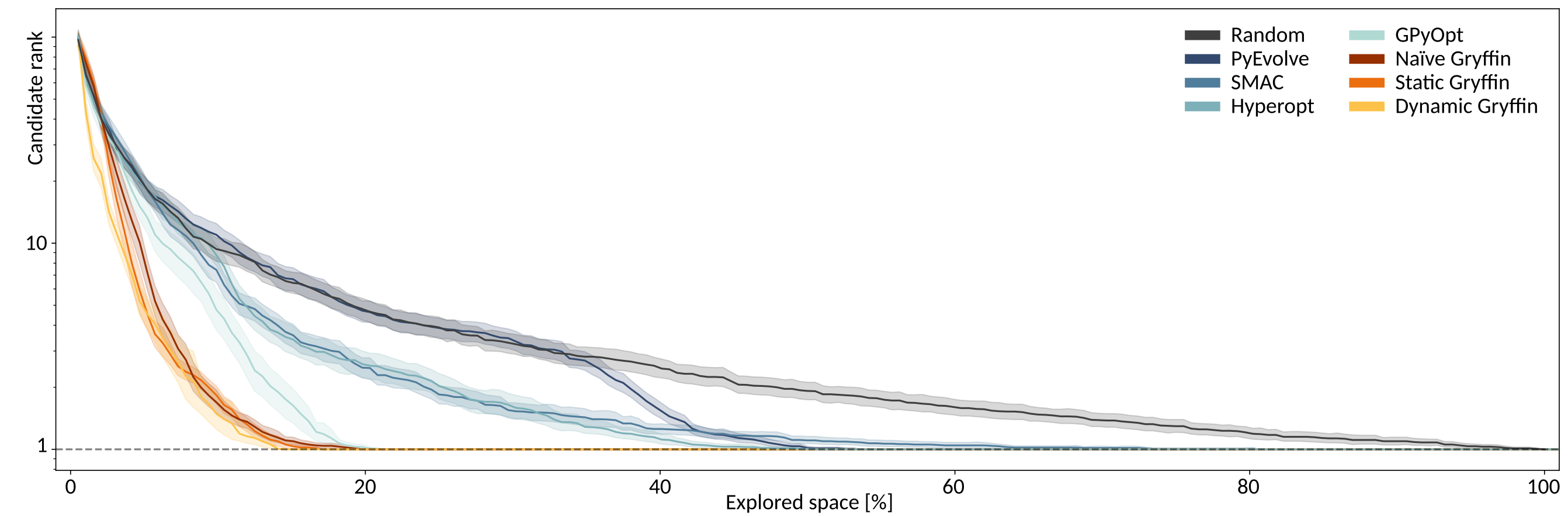}
		\caption{Average rank of the candidate with the lowest bandgap identified by the different optimization strategies during $100$ independent optimization runs.}
		\label{supp_fig:perovskites_results}
	\end{figure}

	Fig.~\ref{supp_fig:perovskites_results} illustrates the performances of the studied optimization strategies on the perovskite application described in detail in Sec.~\ref{sec:perovskites}. Each perovskite candidate is assigned a rank based on its bandgap efficiency, starting from 1 for candidate with the lowest bandgap, and up to 192 for the candidate with the highest bandgap. The graphs shown in Fig.~\ref{supp_fig:perovskites_results} illustrate the average rank of the best performing candidate for different stages of $200$ independent optimization runs.

	In agreement with the results reported in Sec.~\ref{sec:non_fullerenes}, we observe that the dynamic formulation of \gryffin locates promising candidates the fastest. The tendency of static \gryffin to show a reduced optimization rate beyond the initial phase as observed for the non-fullerene application (see Sec.~\ref{supp_sec:non_fullerene_results}) is not as prevalent in this application. 
	
	\subsection{Emulating Suzuki-Miyaura cross-coupling reactions}
\label{supp_sec:suzuki}

	In this application we demonstrate that \gryffin can be used to determine reaction conditions for Suzuki-Miyaura cross-coupling reactions (see Sec.~\ref{sec:suzuki}). Following the experimental procedure detailed by Reizman \emph{et al.}:\cite{Reizman:2016}, we define four controllable reaction parameters: the reaction temperature, the residence time, the catalyst loading, and the ligand forming the Pd complex. These four parameters are selected with the goal to maximize the turn-over number (TON) while maintaining a high yield ($> \unit[85.4]{\%}$). The acceptance threshold for the reaction yield is inspired by the experimental results reported in Ref.~\cite{Reizman:2016} We considered a temperature between \unit[30]{$^\circ$C} and \unit[110]{$^\circ$C}, a residence time between \unit[1]{min} and \unit[10]{min}, and a catalyst loading between \unit[0.5]{\%} and \unit[2.5]{\%}. We further selected one of seven ligands, which are described in more detail in the main text (see Sec.~\ref{sec:suzuki}).

	Repeated executions of individual experiments are avoided by constructing an emulator of the experimental procedure based on previous measurements and using a probabilistic machine learning model. The probabilistic model can reproduce and interpolate measurements obtained from previous experiments, which allows to query the experimental response for any parameter combination via the trained model without the need to run additional experiments. This approach has been demonstrated before across a range of chemistry and materials science applications.\cite{Hase:2018_chimera,Hase:2020_olympus}

	\subsubsection{Constructing an experiment emulator}

		Reizman \emph{et al.}\cite{Reizman:2016} report the reaction yield and TON for a total of $88$ Suzuki-Miyaura cross-coupling reactions,\cite{Reizman:2016} which are used to train a Bayesian neural network (BNN) as a probabilistic model to predict the reaction yield and the TON for any combination of experimental parameters within the search domain. Since BNNs are probabilistic machine learning models, they have the ability to implicitly infer the degree of experimental noise in addition to the expected average response from the presented dataset.

		From the total dataset comprising $88$ reactions, eight reactions were separated for the test set. The reactions for the test set were chosen randomly, but with the constraint that each of the seven ligands was used in at least one of the test set reactions. The remaining $80$ reactions were used for $10$-fold cross-validation. The different ligands were represented as one-hot encoded vectors, and all other experimental conditions were standardized. Both reaction yields and TON were chosen as prediction targets. To account for the fact that both reaction yields and TON cannot be negative, we applied the ReLU activation function to the last layer. Accordingly, the targets were rescaled by dividing by the average reaction yield and TON, respectively, to simplify the initialization of the BNN. We use leaky ReLUs for all other activations and apply dropout at a rate of $0.1$ for further regularization. Distributions for weights and biases are initialized as standard normal distributions. The BNN is constructed with three hidden layers and $24$ neurons per layer. Network parameters are inferred via variational inference using the Adam optimization algorithm with an initial learning rate of $10^{-3}$.

		\begin{figure}[!ht]
			\includegraphics[width = 0.65\textwidth]{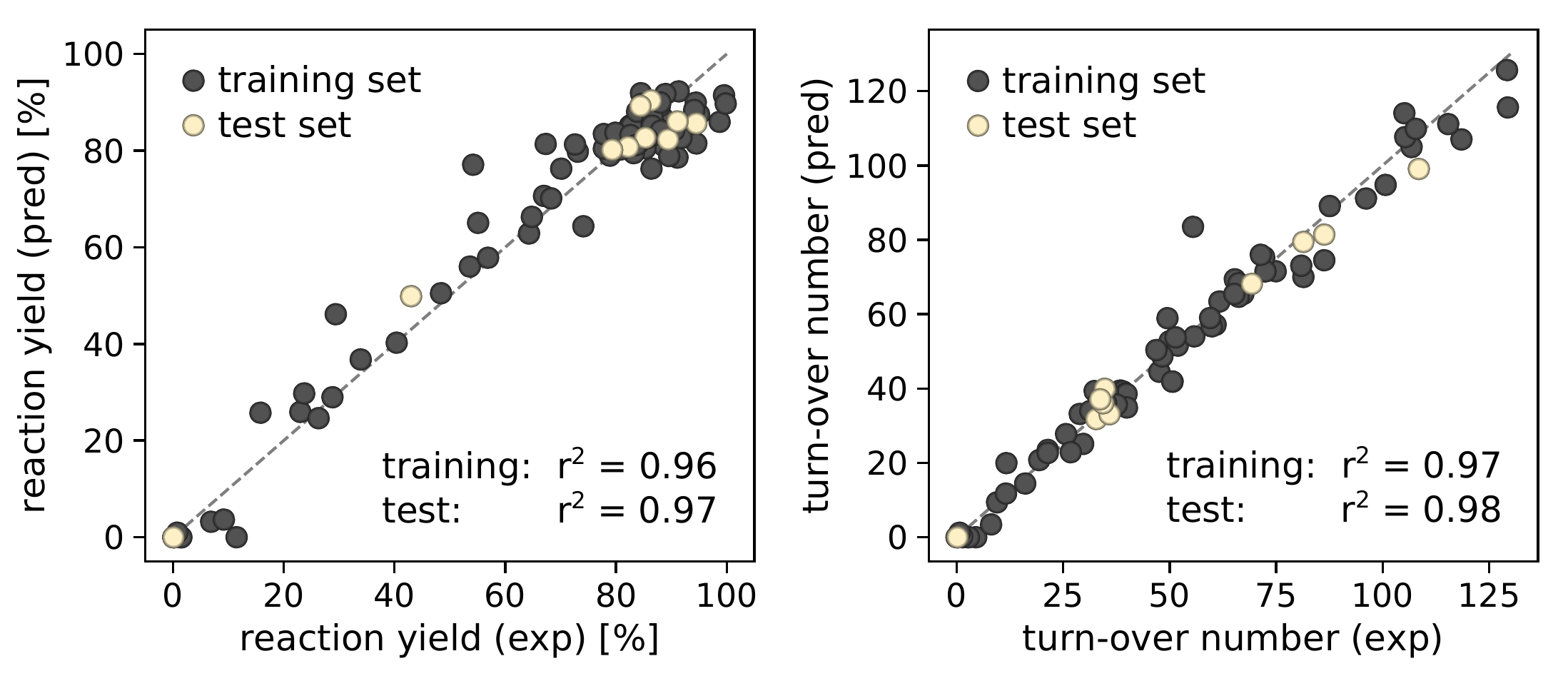}
			\caption{Scatter plot of the emulator predictions for the Suzuki coupling.}
			\label{supp_fig:emulator_accuracies}
		\end{figure}

		Scatter plots of the obtained predictions as well as coefficients of determination are illustrated in Fig.~\ref{supp_fig:emulator_accuracies}. The predicted reactions yields and TONs agree well with the targeted values, which indicates that the trained BNN accurately reproduces the experimental response surfaces of the studied Suzuki-Miyaura reactions.

	\subsubsection{Analysis of ligand descriptors}

\begin{figure}[!ht]
			\centering
			\includegraphics[width = 1.0\textwidth]{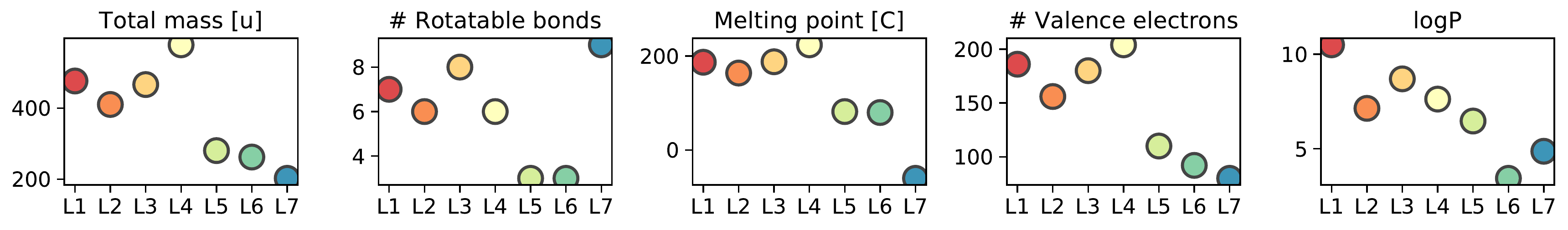}
			\caption{Values of the physical descriptors assigned to the individual ligands (L1-L7) to guide static and dynamic \gryffin for maximization of the reaction yield and the turnover number of the controlled Suzuki reaction. }
			\label{supp_fig:suzuki_descriptors}
		\end{figure}

		The optimization runs with static and dynamic \gryffin are guided by a set of physicochemical descriptors for each of the ligands. Descriptors were chosen based on their availability and computed with RDKit.\cite{rdkit:2006} Descriptor values for each of the ligand choices are illustrated in Fig.~\ref{supp_fig:suzuki_descriptors}

		We compute the yield and TONs for each ligand on a grid of $100$ equidistant levels for each of the three remaining process conditions (temperature, residence time, catalyst loading). The maximum and average values of both optimization objectives, associated to each ligand, are reported in Tab.~\ref{supp_tab:suzuki_dataset1}. We observe that ligand L3 is associated with the highest maximum TON and ligand L1 with the highest maximum yield, while the highest average turnover numbers and yields are achieved with ligand L2. 

		\begin{table}[!ht]
			\centering
			\begin{tabular}{lrrrr}
					 & Max TON & Average TON & Max yield [\%] & Average yield [\%] \\
					 \hline
				L1 & 129.3   & 45.1   & \textbf{96.3}  &  59.7              \\
				L2 & 111.8   & \textbf{46.8} & 95.3         &  \textbf{64.3}     \\
				L3 & \textbf{132.5}   & 45.1        & 92.7           &  59.2              \\
				L4 &  62.4   & 17.7        & 88.1           &  28.3              \\
				L5 &  76.3   &  7.5        & 62.5           &  10.5              \\
				L6 &  47.1   &  3.5        & 39.9           &  5.4               \\
				L7 &  52.6   &  3.3        & 41.2           &  4.9               \\
			\end{tabular}
			\caption{Maximum and average turnover numbers (TON) and reaction yields achieved by each ligand. The highest TONs and yields across all ligands are indicated in bold. }
			\label{supp_tab:suzuki_dataset1}
		\end{table}

		We also analyze the correlation of the provided descriptors with the maximum and average TONs and yields and report the results in Tab.~\ref{supp_tab:suzuki_dataset2}. The logP values of the ligands correlate the best with the maximum TON, the average TON, and the average yield, while the number of valence electrons is most indicative of the maximum achievable yield. Given that the first optimization target is the maximization of the reaction yield, the number of valence electrons is expected to be most informative in the initial phase of the optimization. In addition, because \gryffin will investigate the most promising ligands in more depth, and given that the yield is the primary objective, the number of valence electrons might be considered more informative than logP across the rest of the optimization too.

		\begin{table}[!ht]
			\centering
			\begin{tabular}{lrrrr}
					              & Max TON        & Average TON    & Max yield [\%] & Average yield [\%] \\
					              \hline
			Molecular weight  & 0.527          & 0.649          & 0.873          & 0.685              \\
			Rotatable bonds   & 0.361          & 0.390          & 0.278          & 0.385              \\
			Melting point     & 0.592          & 0.690          & 0.866          & 0.718              \\
			Valence electrons & 0.634          & 0.729          & \textbf{0.923} & 0.759              \\
			logP              & \textbf{0.847} & \textbf{0.805} & 0.892          & \textbf{0.806}     \\
			\end{tabular}
			\caption{Pearson correlation coefficients between individual ligand descriptors and properties of interest. Largest correlations are printed in bold.}
			\label{supp_tab:suzuki_dataset2}
		\end{table}

		Finally, we compute the pairwise correlations between the provided descriptors (Tab.~\ref{supp_tab:desc_correlation}). We find that the number of valence electrons, the melting point and the molecular weight generally correlate well with one another.

		\begin{table}[!ht]
			\centering
			\begin{tabular}{lrrrrr}
				                  & \rotatebox{90}{Molecular weight} & \rotatebox{90}{Rotatable bonds} & \rotatebox{90}{Melting point} & \rotatebox{90}{Valence electrons} & \rotatebox{90}{logP} \\
				Molecular weight  & 1.00 &  0.20 &  0.93 & 0.99 & 0.76 \\
				Rotatable bonds   & 0.20 &  1.00 & -0.08 & 0.26 & 0.39 \\
				Melting point     & 0.93 & -0.08 &  1.00 & 0.93 & 0.70 \\
				Valence electrons & 0.99 &  0.26 &  0.93 & 1.00 & 0.84 \\
				logP              & 0.76 &  0.39 &  0.70 & 0.84 & 1.00 \\
			\end{tabular}
			\caption{Pairwise Pearson correlation between physical descriptors of individual ligands. }
			\label{supp_tab:desc_correlation}
		\end{table}

	\putbib[main]
\end{bibunit}

\end{document}